\PassOptionsToPackage{table}{xcolor}
\documentclass{article}

\usepackage{microtype}
\usepackage{graphicx}
\usepackage{subfigure}
\usepackage{booktabs} 

\usepackage{hyperref}

\usepackage[accepted]{icml2025}

\usepackage{amsmath}
\usepackage{amssymb}
\usepackage{mathtools}
\usepackage{amsthm}

\usepackage{url}
\usepackage{makecell, multirow}
\usepackage{setspace}
\usepackage{bm}
\definecolor{myblue}{RGB}{194, 214, 232}
\definecolor{myred}{RGB}{234, 185, 181}
\definecolor{mycolor}{RGB}{245, 223, 187}
\definecolor{rainbow_red}{RGB}{255, 173, 173}
\definecolor{rainbow_orange}{RGB}{255, 214, 165}
\definecolor{rainbow_yellow}{RGB}{253, 255, 182}
\definecolor{rainbow_green}{RGB}{202, 255, 191}
\definecolor{rainbow_cyan}{RGB}{155, 246, 255}
\definecolor{rainbow_blue}{RGB}{160, 196, 255}
\definecolor{rainbow_purple}{RGB}{189, 178, 255}
\newcommand{\ourmodel}{AEPO}

\newboolean{useBlue}
\setboolean{useBlue}{True} 
\newcommand{\revise}[1]{%
  \ifthenelse{\boolean{useBlue}}%
    {\textcolor{blue}{#1}}
    {\textcolor{black}{#1}}
}
\newcommand{\figureresolution}{low resolution}

\usepackage[capitalize,noabbrev]{cleveref}

\theoremstyle{plain}
\newtheorem{theorem}{Theorem}[section]

\theoremstyle{definition}

\theoremstyle{remark}

\usepackage[textsize=tiny]{todonotes}

\icmltitlerunning{Analytic Energy-Guided Policy Optimization for Offline Reinforcement Learning}

\begin{document}

\twocolumn[
\icmltitle{Analytic Energy-Guided Policy Optimization for Offline Reinforcement Learning}



\icmlsetsymbol{equal}{*}

\begin{icmlauthorlist}
\icmlauthor{Jifeng Hu}{jlu}
\icmlauthor{Sili Huang}{}
\icmlauthor{Zhejian Yang}{}
\icmlauthor{Shengchao Hu}{}
\icmlauthor{Li Shen}{}
\icmlauthor{Hechang Chen}{}
\icmlauthor{Lichao Sun}{}\\
\icmlauthor{Yi Chang}{}
\icmlauthor{Dacheng Tao}{}
\end{icmlauthorlist}

\icmlaffiliation{jlu}{Jilin University}

\icmlcorrespondingauthor{Hechang Chen}{chenhc@jlu.edu.cn}

\icmlkeywords{Machine Learning, ICML}

\vskip 0.3in
]



\printAffiliationsAndNotice{\icmlEqualContribution} 

\begin{abstract}
Conditional decision generation with diffusion models has shown powerful competitiveness in reinforcement learning (RL). 
Recent studies reveal the relation between energy-function-guidance diffusion models and constrained RL problems. 
The main challenge lies in estimating the intermediate energy, which is intractable due to the log-expectation formulation during the generation process.
To address this issue, we propose the Analytic Energy-guided Policy Optimization (AEPO).
Specifically, we first provide a theoretical analysis and the closed-form solution of the intermediate guidance when the diffusion model obeys the conditional Gaussian transformation.
Then, we analyze the posterior Gaussian distribution in the log-expectation formulation and obtain the target estimation of the log-expectation under mild assumptions.
Finally, we train an intermediate energy neural network to approach the target estimation of log-expectation formulation.
We apply our method in 30+ offline RL tasks to demonstrate the effectiveness of our method.
Extensive experiments illustrate that our method surpasses numerous representative baselines in D4RL offline reinforcement learning benchmarks.
\end{abstract}

\section{Introduction}
Controllable generation with diffusion models has shown remarkable success in text-to-image generation~\cite{cao2024controllable}, photorealistic image synthesization~\cite{yang2024improving}, high-resolution video creation~\cite{videoworldsimulators2024}, and robotics manipulation~\cite{chen2024simple}.
A common strategy to realize controllable diffusion models is guided sampling, which can be further classified into two categories: classifier-guided~\cite{dhariwal2021diffusion, janner2022planning} and classifier-free-guided~\cite{liu2023more, ajay2022conditional} diffusion-based methods.

Usually, classifier guidance and classifier-free guidance need paired data, i.e., samples and the corresponding conditioning variables, to train a controllable diffusion model~\cite{yang2024using, graikos2022diffusion, rombach2022high}.
However, it is difficult to describe the conditioning variables for each transition in RL.
We can only evaluate the value for the transitions with a scalar and continuous function, which is generally called the action value function (Q function)~\cite{lu2023contrastive, janner2022planning}.
Recently, energy-function-guided diffusion models provide an effective way to realize elaborate manipulation in RL.
Previous studies also reveal the relation of equivalence between guided sampling energy-function-guided diffusion models and constrained RL problems~\cite{lu2023contrastive, chen2022offline}.
Considering the following formula of energy-function-guided diffusion models 
\begin{equation}\label{energy-function-guided diffusion models}
    \begin{aligned}
    \min_{p}~~&\mathbb{E}_{x\sim p(x)}\mathcal{E}(x),\\
        s.t.~~&D_{KL}(p(x)||q(x)),\\
    \end{aligned}
\end{equation}
where $x\in\mathbb{R}^d$, $\mathcal{E}(x):\mathbb{R}^d\rightarrow\mathbb{R}$ is the energy function, $q(x)$ and $p(x)$ denote the unguided and guided data distribution, and $D_{KL}(\cdot)$ represents the KL divergence. 
The optimal solution to the above problem is 
\begin{equation}\label{solution of energy-function-guided diffusion models}
    p(x)\propto q(x)e^{-\beta\mathcal{E}(x)},
\end{equation}
where inverse temperature $\beta$ is the Lagrangian multiplier, which controls the energy strength. 
Equation~\eqref{solution of energy-function-guided diffusion models} shows that the guided data distribution lies in the intersection region of unguided distribution $q(x)$ and energy distribution $e^{-\beta\mathcal{E}(x)}$.
Obversely, the guidance comes from the energy function $\mathcal{E}(x)$.
The constrained RL problem 
\begin{equation}\label{constrained RL problem}
    \begin{aligned}
        \max_{\pi}~~&\mathbb{E}_{s\sim D^\mu}\left[\mathbb{E}_{a\sim\pi(\cdot|s)}Q(s,a)\right],\\
        s.t.~~&D_{KL}(\pi(\cdot|s)||\mu(\cdot|s)),\\
    \end{aligned}
\end{equation}
has the similar formula with Equation~\eqref{energy-function-guided diffusion models},
where $\pi$ is the learned policy, and $\mu$ is the behavior policy.
Thus,  
\begin{equation}\label{solution of constrained RL problem}
    \pi^*(a|s)\propto\mu(a|s)e^{\beta Q(s,a)}.
\end{equation}
The optimal policy $\pi^*(a|s)$ under state $s$ can be regarded as the guided distribution $p(x)$, and actions $a$ can be generated with guided sampling.
See Appendix~\ref{The Constrained RL Problem} for more details.

Although sampling from $p(x)$ is intractable due to the normalization term, it can be bypassed in score-based diffusion models, where the score function $\nabla_x \log~p(x)$ can be calculated with intermediate energy $\mathcal{E}_t(x_t)$ (Refer to Equation~\eqref{general intermediate guidance} for details.)~\cite{liu2023more, ho2022classifier}.
However, the intermediate energy introduced in classifier-guided and classifier-free-guided methods has proved to be inexact theoretically and practically~\cite{lu2023contrastive, yuan2024reward}.
Besides, for classifier-guided methods, the intermediate guidance can be influenced by the fictitious state-action pairs generated during the generation process. 
For classifier-free-guided methods, the intermediate guidance is manually predefined during guided sampling.
Though recent studies~\cite{lu2023contrastive} propose the exact expression of $\mathcal{E}_t(x_t)$, they do not investigate the theoretical solution.

In this paper, we theoretically analyze the intermediate energy $\mathcal{E}_t(x_t)$ (intermediate guidance is $\nabla_{x_t}\mathcal{E}_t(x_t)$) with a log-expectation formulation and derive the solution of intermediate energy by addressing the posterior integral.
Based on the theoretical results, we propose a new diffusion-based method called Analytic Energy-guided Policy Optimization (\ourmodel{}).
In contrast with previous studies, we leverage the characteristic of Gaussian distribution and find a solution for the log-expectation formulation.
Specifically, we first convert the implicit dependence on the action in the exponential term to explicit dependence by applying Taylor expansion.
After that, we investigate the posterior distribution formulation of the expectation term and simplify the intractable log-expectation formulation with the moment-generating function of the Gaussian distribution.
Then, we propose to train a general Q function and an intermediate energy function to approach the simplified log-expectation formulation given a batch of data.
Finally, during the inference, we can use the intermediate guidance to generate guided information to high-return action distributions.

To verify the effectiveness of our method, we apply our method in offline RL benchmarks D4RL~\cite{fu2020d4rl}, where we select different tasks with various difficulties.
We compare our method with dozens of baselines, which contain many types of methods, such as classifier-guided and classifier-free-guided diffusion models, behavior cloning, and transformer-based models.
Through extensive experiments, we demonstrate that our method surpasses state-of-the-art algorithms in most environments.
\section{Preliminary}

\subsection{Diffusion Probabilistic Models}
Diffusion Probabilistic Models (DPMs)~\cite{ho2020denoising, dhariwal2021diffusion, rombach2022high} are proposed to construct the transformation between complex data distribution $x_0$ and easy sampling distribution $x_T$ (e.g., Gaussian distribution).
By defining the forward transformation from data distribution $x_0$ to simple distribution $x_T$ in $T$ time range according to the following stochastic differential equation
\begin{equation*}
    dx = f(x, t)dt+g(t) d\text{w},
\end{equation*}
where $f(\cdot, t):\mathbb{R}^d\rightarrow\mathbb{R}^d$ is the drift coefficient, $g(\cdot)$ is the diffusion coefficient, and w is the standard Wiener process, we can obtain a reverse transformation from the simple distribution $x_T$ to data distribution $x_0$, as shown in 
\begin{equation*}
    dx = \left[f(x, t)-g(t)^2\nabla_{x} \log~q_t(x)\right]dt + g(t)d\bar{\text{w}},
\end{equation*}
where $\bar{\text{w}}$ is the standard Wiener process, $t\in[0,T]$ is the diffusion time, $t=0$ means data without perturbation, $t=T$ means prior sample-friendly distribution, $dt$ indicates infinitesimal negative timestep, $q_t(x)$ is the marginal distribution of $x_t$, and $\nabla_{x} \log~q_t(x)$ is the score function. 
Usually, we adopt the Gaussian transition distribution~\cite{lu2022dpm, lu2022dpmb}, which means
\begin{equation}\label{conditional gaussian on x0}
    q_{t|0}(x_t|x_0) = \mathcal{N}(x_t;\alpha_t x_0, \sigma_t^2\bm{I}),
\end{equation}
where $\alpha_t>0$ is a decrease function, $\sigma_t>0$ is an increase function, $q_{T|0}(x_T|x_0)\approx\mathcal{N}(x_T;0, \tilde{\sigma}^2\bm{I})$ for certain $\tilde{\sigma}$.
Obviously, $q_{T|0}(x_T|x_0)$ will be independent of $x_0$, i.e., $q_{T|0}(x_T|x_0)\approx q_{T}(x_T)$.
Based on the above results, once we obtain $\nabla_{x} \log~q_t(x)$, we can perform generation from simple distribution~\cite{bao2022estimating, yang2023diffusion}.
By constructing the score function $s(x_t)$, previous studies~\cite{song2020score, song2019generative} show that the following objective is equivalent:
\begin{equation*}
    \begin{aligned}
        &\mathbb{E}_{q(x_t)}\left[||s(x_t)-\nabla_{x_t} \log~q_t(x_t)||^2_2\right]\\
        \Rightarrow&\mathbb{E}_{ q(x_0)q_{t|0}(x_t|x_0)}\left[||s(x_t)-\nabla_{x_t} \log~q_{t|0}(x_t|x_0)||^2_2\right]
    \end{aligned}
\end{equation*}
and the reverse transformation can be an alternative formula, i.e., probability flow ordinary differential equation (ODE) 
\begin{equation}\label{neural ODE}
    \frac{dx_t}{dt}=f(t)x_t-\frac{1}{2}g(t)^2\nabla_{x_t}\log~p_t(x_t),
\end{equation}
where $f(x,t)=f(t)x_t$, $f(t)=\frac{d~\log\alpha_t}{dt}$, and $g(t)^2=\frac{d~\sigma_t^2}{dt}-2\frac{d~\log\alpha_t}{dt}\sigma_t^2$, because Equation~\eqref{neural ODE} holds the marginal distribution unaltered~\cite{kingma2021variational}.
Considering that $x_t=\alpha_tx_0+\sigma_t\epsilon$, where $\epsilon\sim\mathcal{N}(0,\bm{I})$, there exist $s(x_t)=\nabla_{x_t}\log~p_t(x_t)\approx -\frac{\epsilon}{\sigma_t}$.
So, we can introduce a neural network $\epsilon_\theta(x_t, t)$, and the diffusion loss is 
\begin{equation}\label{general diffusion loss}
    \mathcal{L}_{diff}=\mathbb{E}_{x_0\sim q(x_0),t\sim U(0,T)}\left[||\epsilon_\theta(x_t,t)-\epsilon||_2^2\right],
\end{equation}
where $U(\cdot)$ is uniform distribution.
$x_0\sim q(x_0)$ means sampling data from the offline datasets.
After training, we can obtain the score function value through $\epsilon_\theta$, i.e., $\nabla_{x_t}\log~p_t(x_t)\approx -\frac{\epsilon_\theta}{\sigma_t}$.
Then, we can use Equation~\eqref{neural ODE} for the generation process~\cite{lu2022dpm}.

\subsection{Guided Sampling}
In RL, guidance plays an important role in generating plausible decisions because the dataset quality is usually mixed, and naive modeling of the conditional action distribution under states will lead to suboptimal performance.
Classifier-guided methods~\cite{dhariwal2021diffusion, wang2022diffusion, janner2022planning, kang2024efficient} define binary random optimality variables $\mathcal{O}$, where $\mathcal{O}=1$ means optimal outputs, and $\mathcal{O}=0$ means suboptimal outputs.
For each generation step $t$, $p(x_{t-1}|x_t,\mathcal{O})\propto q(x_{t-1}|x_t)p(\mathcal{O}|x_t)$, the guidance serves as a gradient on the mean value modification $p(x_{t-1}|x_{t}, \mathcal{O})=\mathcal{N}(x_t; \mu_t+\Sigma_t\cdot\nabla \log~p(\mathcal{O}|x_t), \Sigma_t)$, where $\mu_t$ and $\Sigma_t$ is the predicted mean and variances of $q(x_{t-1}|x_{t})$.
During inference, the fictitious intermediate outputs $x_t$ that do not exist in the training dataset will lead to suboptimal intermediate guidance.
Different from classifier-guided methods, the classifier-free-guided methods implicitly build the joint distribution between the data and condition variables $\mathcal{C}$ in the training phase~\cite{ajay2022conditional, chen2022offline, liu2023more, chen2023score}.
Therefore, we can use the desired condition as guidance to perform guided sampling. 
The classifier-free-guided training loss is 
$\mathbb{E}_{x_0\sim q(x_0),t\sim U(0,T),b\sim\mathcal{B}(\lambda)}\left[||\epsilon_\theta(x_t,b*\mathcal{C},t)-\epsilon||_2^2\right]$,
where $\mathcal{B}$ is binomial distribution, $\lambda$ is the parameter of $\mathcal{B}$.
During inference, the guidance is also incorporated in the mixed prediction of conditional and unconditional noise, i.e.,
$\hat{\epsilon}=\epsilon_{\theta}(x_t, t, \emptyset)+\omega(\epsilon_{\theta}(x_t, t, \mathcal{C})-\epsilon_{\theta}(x_t, t, \emptyset))$, $\omega$ controls the guidance strength, $\emptyset$ means $b=0$.
Due to the condition variables being needed before generation, we need to manually assign the value before inference.
The predefined $\mathcal{C}$ will also restrict the model's performance if insufficient prior information on environments arises.

\subsection{Diffusion Offline RL}
Typical RL~\cite{mnih2013playing} is formulated by the Markov Decision Process (MDP) that is defined as the tuple $\mathcal{M}=\langle\mathcal{S}, \mathcal{A}, \mathcal{P}, r, \gamma\rangle$, where $\mathcal{S}$ and $\mathcal{A}$ denote the state and action space, respectively, $\mathcal{P}(s^\prime|s,a)$ is the Markovian transition probability, $r(s,a)$ is the reward function, and $\gamma\in [0, 1)$ is the discount factor.
The goal is to find a policy $\pi$ that can maximize the discounted return $\mathbb{E}_{\pi}[\sum_{k=0}^{\infty}\gamma^{k}r(s_k, a_k)]$, where $k$ represents the RL time step which is different from diffusion step $t$~\cite{schulman2017proximal}.
In offline RL~\cite{kumar2020conservative, kostrikov2021offline}, only a static dataset $D_{\mu}$ collected with behavior policy is available for training.
Extracting optimal policy from offline datasets is formulated as a constrained RL problem~\cite{peng2019advantage, wu2019behavior} as shown in Equation~\eqref{constrained RL problem}, which can be converted to find an optimal policy $\pi^*$ that maximizes 
$\pi^*=\max_{\pi}\mathbb{E}_{s\sim D_\mu}\mathbb{E}_{a\sim\pi(\cdot|s)}\left[Q(s,a)-\frac{1}{\beta}D_{KL}(\pi(\cdot|s)||\mu(\cdot|s))\right]$.
Diffusion offline RL usually adopts diffusion models to imitate the behavior policy $\mu$~\cite{ajay2022conditional}. 
Naively modeling the action distribution will only lead to suboptimal policy.
Thus, the well-trained value functions will be used to extract optimal policy, such as 
action value gradient guidance~\cite{janner2022planning} and empirical action distribution reconstruction~\cite{hansen2023idql}.

\section{Method}\label{Method}

In order to generate samples from the desired distribution $p(x)$ with the reverse transformation (generation process) Equation~\eqref{neural ODE}, we should know the score function $\nabla_x \log~p(x)$.
If the score function of the desired distribution $p(x)$ has a relation with the score function of $q(x)$ 
\begin{equation}\label{general intermediate guidance}
    \nabla_x \log~p(x) = \nabla_x \log~q(x) +\nabla_x -\beta\mathcal{E}(x),
\end{equation}
for any data $x$, we can obtain the score function of $p(x)$ by compounding the score function of $q(x)$ and the gradient of $\mathcal{E}(x)$.
However, it only exists $p_0(x_0)\propto q_0(x_0)e^{-\beta\mathcal{E}(x_0)}$ for the samples in the dataset rather than the marginal distribution of $p_t(x_t)$ and $q_t(x_t)$.
Previous guided sampling methods usually adopt MSE or diffusion posterior sampling (DPS) as the objective of training the intermediate energy, which cannot satisfy the relation $p_t(x_t)\propto q_t(x_t)e^{-\mathcal{E}_t(x_t)}$, thus leading to an inexact intermediate guidance.
We summarize the results in Theorem~\ref{Inexact and Exact Intermediate Energy}.

\begin{theorem}[Inexact and Exact Intermediate Energy]
\label{Inexact and Exact Intermediate Energy}
Suppose $p_0(x_0)$ and $q_0(x_0)$ has the relation of Equation~\eqref{solution of energy-function-guided diffusion models}. $p_{t|0}(x_t|x_0)$ and $q_{t|0}(x_t|x_0)$ are defined by
\begin{equation}\label{marginal distribution of p and q}
    p_{t|0}(x_t|x_0) \coloneqq q_{t|0}(x_t|x_0) = \mathcal{N}(x_t;\alpha_t x_0,\sigma_t^2\bm{I})
\end{equation}
for all $t\in(0, T]$.
According to the Law of Total Probability, the marginal distribution $p_t(x_t)$ and $q_t(x_t)$ are given by 
$p_t(x_t)=\int p_{t|0}(x_t|x_0) p_0(x_0)dx_0$ and $q_t(x_t)=\int q_{t|0}(x_t|x_0) q_0(x_0)dx_0$.
Previous studies~\cite{janner2022planning, chung2022diffusion} define the inexact intermediate energy as 
\begin{equation}\label{inexact intermediate energy when non zero t}
    \begin{aligned}
        \mathcal{E}^{MSE}_t(x_t)&=\mathbb{E}_{q_{0|t}(x_0|x_t)}[\mathcal{E}(x_0)], t>0,\\
        \mathcal{E}^{DPS}_t(x_t)&=\mathcal{E}(\mathbb{E}_{q_{0|t}(x_0|x_t)}[x_0]), t>0.
    \end{aligned}
\end{equation}
The exact intermediate energy is defined by 
\begin{equation}\label{exact intermediate energy when non zero t}
    \mathcal{E}_t(x_t)=-\log~\mathbb{E}_{q_{0|t}(x_0|x_t)}[e^{-\beta\mathcal{E}(x_0)}], t>0.
\end{equation}
It can be proved (See Appendix~\ref{Exact Intermediate Guidance} for details.) that $p_t(x_t) \propto q_t(x_t)e^{-\mathcal{E}_t(x_t)}$
exists under Equation~\eqref{exact intermediate energy when non zero t} rather than Equation~\eqref{inexact intermediate energy when non zero t}.
So, the intermediate guidance (Equation~\eqref{general intermediate guidance}) is inexact for previous classifier-guided and classifier-free-guided methods.
\end{theorem}

Similarly, in RL, as shown in Equation~\eqref{solution of constrained RL problem}, the desired distribution is $\pi(a|s)$. The intermediate energy $\mathcal{E}_t(s,a_t)$ and the score function $\nabla_{a_t}\log \pi_t(a_t|s)$ with intermediate guidance $\nabla_{a_t}\mathcal{E}_t(s,a_t)$ in RL are defined as follows
\begin{equation}\label{intermediate energy definition in RL}
    \mathcal{E}_t(s,a_t)=
    \begin{cases}
    \beta Q(s,a_0), & t = 0 \\
    \log~\mathbb{E}_{\mu_{0|t}(a_0|a_t,s)}[e^{\beta Q(s,a_0)}], & t > 0
    \end{cases}
\end{equation}
\begin{equation}\label{intermediate energy_guidance in RL}
    \nabla_{a_t}\log \pi_t(a_t|s) = \nabla_{a_t} \log~\mu_t(a_t|s) + \nabla_{a_t} \mathcal{E}_t(s, a_t),
\end{equation}
where we slightly abuse the input of $\mathcal{E}$ because the value action should depend on the corresponding state in RL.
Refer to Appendix~\ref{Guidance of Intermediate Diffusion Steps} for the detailed derivation.

\subsection{Intermediate Energy}

Diffusion loss (Equation~\eqref{general diffusion loss}) provides the way to obtain the score function of $\mu_t(a_t|s)$, 
\begin{equation}\label{unguided intermediate guidance}
    \nabla_{a_t}\log~\mu_t(a_t|s) = -\frac{\epsilon_\theta(s, a_t, t)}{\sigma_t}.
\end{equation}
Obviously, the most challenging issue that we need to address is $\log~\mathbb{E}_{\mu_{0|t}(a_0|a_t,s)}[e^{\beta Q(s,a_0)}]$ because of the intractable log-expectation formulation.
Reminding that the posterior distribution $\mu_{0|t}(a_0|a_t,s)$ is also a Gaussian distribution, where the mean and covariance are denoted as $\mu_{0|t}=\mathcal{N}(\tilde{\mu}_{0|t},\tilde{\Sigma}_{0|t})$.
We will introduce how to approximate $\tilde{\mu}_{0|t}$ and $\tilde{\Sigma}_{0|t}$ in next section.
Now, we first focus on converting the implicit dependence on the action in the exponential term to explicit dependence by applying Taylor expansion.
Expand $Q(s, a_0)$ at $a_0=\bar{a}$ with Taylor expansion, where $\bar{a}$ is a constant vector.
Then, we have
\begin{equation}\label{taylor expansion of Q}
    \begin{aligned}
        Q(s, a_0) &\approx Q(s, a_0)|_{a_0=\bar{a}} + \frac{\partial Q(s, a_0)}{\partial a_0}^\top|_{a_0=\bar{a}}*(a_0-\bar{a}).
    \end{aligned}
\end{equation}
Replacing Q function in $\log~\mathbb{E}_{a_{0}\sim \mu(a_0|a_t,s)}e^{\beta Q(s, a_0)}$ with Equation~\eqref{taylor expansion of Q}, we will derive the following approximation
\begin{equation}\label{moment generating function}
    \begin{aligned}
        &\log~\mathbb{E}_{a_{0}\sim \mu(a_0|a_t,s)}e^{\beta Q(s, a_0)}\\
        \approx&\log~\mathbb{E}_{a_{0}\sim \mu(a_0|a_t,s)}e^{\beta\left(Q(s, a_0)|_{a_0=\bar{a}} + \frac{\partial Q(s, a_0)}{\partial a_0}^\top|_{a_0=\bar{a}}*(a_0-\bar{a})\right)}\\
        =&\beta Q(s, \bar{a})-\beta Q^{\prime}(s,\bar{a})^\top \bar{a}\\
        &+\log~\left\{\mathbb{E}_{a_{0}\sim \mu(a_0|a_t,s)}[e^{\beta Q^{\prime}(s, \bar{a})^\top a_0}]\right\},
    \end{aligned}
\end{equation}
where $Q^{\prime}=\frac{\partial Q}{\partial a}$.
Note that the above derivation makes the complex dependence between $Q(s, a_0)$ and $a_0$ easier.
In other words, we transfer $a_0$ from the implicit dependence on $Q(s, a_0)$ to explicit dependence on $Q^{\prime}(s, \bar{a})^\top a_0$.
Refer to Appendix~\ref{Detailed Derivation of Intermediate Guidance} for more details.
As for the only unknown item $\mathbb{E}_{a_{0}\sim \mu(a_0|a_t,s)}[e^{\beta Q^{\prime}(s, \bar{a})^\top a_0}]$, the exact result can be derivated from moment generating function, which indicates that $\mathbb{E}_{x\sim\mathcal{N}(v, \Sigma)}[e^{a^{\top}x}] = e^{a^\top v+\frac{1}{2}a^\top\Sigma a}$.
Applying the above results to Equation~\eqref{moment generating function}, we have 
\begin{equation*}
    \begin{aligned}
        \log~&\left\{\mathbb{E}_{a_{0}\sim \mu(a_0|a_t,s)}[e^{\beta Q^{\prime}(s, \bar{a})^\top a_0}]\right\}\\
        &=\beta Q^{\prime}(s, \bar{a})^\top \tilde{\mu}_{0|t}+\frac{1}{2} \beta^2 Q^{\prime}(s, \bar{a})^\top\tilde{\Sigma}_{0|t}Q^{\prime}(s, \bar{a}).
    \end{aligned}
\end{equation*}
Finally, the intermediate energy $\mathcal{E}_t(s, a_t)$ with log-expectation formulation can be approximated by 
\begin{equation}\label{intermediate energy approximation}
    \begin{aligned}
        &\log~\mathbb{E}_{a_{0}\sim \mu(a_0|a_t,s)}e^{\beta Q(s, a_0)}\approx\beta Q(s, \bar{a})\\&+\beta Q^{\prime}(s, \bar{a})^\top (\tilde{\mu}_{0|t}-\bar{a})+\frac{1}{2} \beta^2 Q^{\prime}(s, \bar{a})^\top\tilde{\Sigma}_{0|t}Q^{\prime}(s, \bar{a}).\\
    \end{aligned}
\end{equation} 
From Equation~\eqref{intermediate energy approximation}, we can see that the intermediate energy can be approximated with $\bar{a}$, $\tilde{\mu}_{0|t}$, and $\tilde{\Sigma}_{0|t}$.
In the next section, we will introduce how to approximate the parameters of the posterior distribution $\mu(a_0|a_t,s)$.
The training algorithm is shown in Algorithm~\ref{algorithm} of Appendix~\ref{Pseudocode}.

\subsection{Posterior Approximation}\label{Posterior Approximation}
In this paper, we provide several methods to approximate the distribution of $\mu(a_0|a_t,s)$.
In order to distinguish different posterior approximation methods, we use ``Posterior i'' to differentiate them in the following contents. 

\textbf{Posterior 1.}~~~
Inspired by previous studies~\cite{bao2022analytic}, we can use the trained diffusion model $\epsilon_{\theta}$ to obtain the mean vector $\tilde{\mu}_{0|t}$ of the distribution $\mu_{0|t}(a_0|a_t,s) = \mathcal{N}(a_0; \tilde{\mu}_{0|t}, \tilde{\Sigma}_{0|t})$,
where the mean vector 
\begin{equation}\label{posterior mean}
    \tilde{\mu}_{0|t}=\frac{1}{\alpha_t}(a_t-\sigma_t\epsilon_\theta(s,a_t,t))
\end{equation}
according to Equation~\eqref{conditional gaussian on x0}.
As for covariance matrix $\tilde{\Sigma}_{0|t}$, following the definition of covariance
\begin{equation}\label{q0|t analytic expression of posterior 1}
    \begin{aligned}
        &\tilde{\Sigma}_{0|t}(a_t) = \mathbb{E}_{\mu_{0|t}(a_0|a_t,s)}\left[(a_0 - \tilde{\mu}_{0|t})(a_0 - \tilde{\mu}_{0|t})^\top\right]\\
        &=\frac{1}{\alpha_t^2}\mathbb{E}_{\mu_{0|t}(a_0|a_t,s)}\left[(a_t - \alpha_t a_0)(a_t - \alpha_t a_0)^\top\right]-\frac{\sigma_t^2}{\alpha_t^2}\epsilon_\theta\epsilon_\theta^\top,
    \end{aligned}
\end{equation}
where in the second equation we use $\mathbb{E}_{\mu_{0|t}(a_0|a_t)}[(a_0 - \frac{1}{\alpha_t}a_t)*\frac{\sigma_t}{\alpha_t}\epsilon_\theta]=(\mathbb{E}_{\mu_{0|t}(a_0|a_t,s)}[a_0] - \frac{1}{\alpha_t}a_t)*\frac{\sigma_t}{\alpha_t}\epsilon_\theta=(\tilde{\mu}_{0|t} - \frac{1}{\alpha_t}a_t)*\frac{\sigma_t}{\alpha_t}\epsilon_\theta=-\frac{\sigma_t^2}{\alpha_t^2}\epsilon_\theta\epsilon_\theta^\top$, $\epsilon_\theta=\epsilon_\theta(s,a_t,t)$, and $\tilde{\mu}_{0|t}=\mathbb{E}_{\mu_{0|t}(a_0|a_t,s)}[a_0]$.
Besides, we also notice that $\mu(a_t)\sim\mathcal{N}(a_t;\alpha_t a_0,\sigma_t^2\bm{I})$, thus we have
\begin{equation}
    \begin{aligned}
        &\mathbb{E}_{\mu_t(a_t|s)}\mathbb{E}_{\mu_{0|t}(a_0|a_t,s)}\left[(a_t - \alpha_t a_0)(a_t - \alpha_t a_0)^\top\right]=\sigma_t^2\bm{I}.\\
    \end{aligned}
\end{equation}
To make the variance independent from the data $a_t$, we have
\begin{equation}
    \begin{aligned}
        \tilde{\Sigma}_{0|t}&=\mathbb{E}_{\mu_t(a_t|s)}\tilde{\Sigma}_{0|t}(a_t)=\frac{\sigma_t^2}{\alpha^2_t}[\bm{I}-\mathbb{E}_{\mu_t(a_t|s)}[\epsilon_\theta\epsilon_\theta^\top]].
    \end{aligned}
\end{equation}
For simplicity, we usually consider isotropic Gaussian distribution, where the  covariance $\tilde{\Sigma}_{0|t}$ satisfies $\tilde{\Sigma}_{0|t}=\tilde{\sigma}_{0|t}^2 * \bm{I}$, the covariance can be simplified as 
\begin{equation}\label{variance of posterior 1}
    \tilde{\sigma}_{0|t}^2 = \frac{\sigma_t^2}{\alpha^2_t}\left[1-\frac{1}{d}\mathbb{E}_{\mu_t(a_t|s)}\left[||\epsilon_\theta(a_t, t)||^2_2\right]\right].
\end{equation}
In the experiments, we adopt this posterior as the default setting to conduct all experiments.
Refer to Appendix~\ref{Posterior 1 of appendix} for the detailed derivation.

\textbf{Posterior 2.}~~~
As Equation~\eqref{posterior mean} shows, we still adopt the same mean of the distribution $\mu_{0|t}(a_0|a_t,s)$.
We assume that the $q_{0|t}(a_0|a_t)$ obey the formulation $\mu_{0|t}(a_0|a_t,s)=\mathcal{N}(a_0;\tilde{\mu}_{0|t},\tilde{\Sigma}_{0|t})$.
Again considering Equation~\eqref{q0|t analytic expression of posterior 1}, we reformulate~\cite{kexuefm9246} it as 
\begin{equation}
    \begin{aligned}
        \tilde{\Sigma}_{0|t}(a_t)
        &=\mathbb{E}_{\mu_{0|t}(a_0|a_t,s)}\left[(a_0 - u_0)(a_0 - u_0)^\top\right]\\&~~~~-(\tilde{\mu}_{0|t} - u_0)(\tilde{\mu}_{0|t} - u_0)^\top,
    \end{aligned}
\end{equation}
where $u_0$ is any constant vector, which has the same dimension with $\tilde{\mu}_{0|t}$.
When we consider isotropic Gaussian distribution and remove the inference of $a_t$, the $\tilde{\Sigma}_{0|t}$ is simplified as $\tilde{\Sigma}_{0|t}=\tilde{\sigma}_{0|t}^2\bm{I}$ and
\begin{equation}\label{variance of posterior 2}
    \tilde{\sigma}_{0|t}^2 = Var(a_0) - \frac{1}{d}\mathbb{E}_{\mu_t(a_t|s)}[||\tilde{\mu}_{0|t} - u_0||^2_2],
\end{equation}
where $u_0=\mathbb{E}_{a_0\sim\mu(a_0)}[a_0]$ and we can use the mean value of the whole action vectors of the dataset as an unbiased estimation.
$Var(a_0)$ can be approximated from a batch of data or from the entire dataset.
We can sample a batch of data $a_t$ with different $t$ to calculate an approximation solution of the second term $\mathbb{E}_{\mu_t(a_t|s)}[||\tilde{\mu}_{0|t} - u_0||^2_2]$.
In Appendix~\ref{Posterior 2 of appendix}, we provide the detailed derivation.

From the above theory, we can solve the Gaussian distribution $\mu_{0|t}(a_0|a_t,s)$'s parameters $(\tilde{\mu}_{0|t}, \tilde{\Sigma}_{0|t})$.
According to Equation~\eqref{intermediate energy approximation}, we have simplified intermediate energy:
\begin{equation}\label{intermediate energy approximation with IGD}
    \begin{aligned}
        &\mathcal{E}_t(s,a_t)=\log~\mathbb{E}_{a_{0}\sim \mu(a_0|a_t,s)}e^{\beta Q(s, a_0)}\approx\beta Q(s, \bar{a})\\&+\beta Q^{\prime}(s, \bar{a})^\top (\tilde{\mu}_{0|t}-\bar{a})+\frac{1}{2} \beta^2 \tilde{\sigma}_{0|t}^2*||Q^{\prime}(s, \bar{a})||^2_2,\\
    \end{aligned}
\end{equation}
and the intermediate energy training loss $\mathcal{L}_{IE}$ is 
\begin{equation}\label{intermediate energy loss}
    \mathcal{L}_{IE}=\mathbb{E}\left[||\mathcal{E}_\Theta(s,a_t,t)-\mathcal{E}_t(s,a_t)||_2^2\right],
\end{equation}
where $\Theta$ is the parameter of intermediate energy, the mean and covariance are given by Equation~\eqref{posterior mean},~\eqref{variance of posterior 1}, and~\eqref{variance of posterior 2}.

\subsection{Q Function Training}
Due to the fact that $\mathcal{E}_t(s,a_t)$ rely on $Q(s, a)$, we first incorporate an action value function $Q_{\psi}(s, a)$ with parameters $\psi$ to approximate the action values.
As for the training of the $Q_{\psi}(s,a)$, we leverage the expectile regression loss to train the Q function and V function:
\begin{equation}\label{critic functions training loss}
    \begin{aligned}
        &\mathcal{L}_{V}=\mathbb{E}_{(s,a)\sim D_\mu}\left[L_2^\tau(V_{\phi}(s)-Q_{\bar{\psi}}(s,a))\right],\\
        &\mathcal{L}_{Q}=\mathbb{E}_{(s,a,s^\prime)\sim D_\mu}\left[||r(s,a) + V_{\phi}(s^\prime)-Q_{\psi}(s,a)||^2_2\right],\\
        &L_2^\tau(y)=|\tau-1(y<0)|y^2,
    \end{aligned}
\end{equation}
where $\phi$ is the parameters of value function $V$, $\bar{\psi}$ is the parameters of target Q, $\tau$ controls the weights of different $y$.
There are also some other methods suitable for learning the Q function from offline datasets, such as In-support Q-learning~\cite{lu2023contrastive} and conservative Q-learning~\cite{kumar2020conservative}.
However, they use either fake actions that are generated from generative models or over-underestimate values for out-of-dataset actions, which influence the learned Q values and the calculation of the intermediate energy.
We defer more discussion of the training of the Q function on offline datasets in Appendix~\ref{The Training of Q Function}.

\subsection{Guidance Rescaling}
 
As shown in Equation~\eqref{intermediate energy_guidance in RL}, the magnitude and direction of $\nabla_{a_t}\log~\pi_t(a_t|s)$ will be easily affected by the gradient of the intermediate energy.
Previous studies usually need extra hyperparameter $w$ to adjust the guidance degree to find better performance, i.e., $\nabla_{a_t}\log \pi_t(a_t|s) = \nabla_{a_t} \log~\mu_t(a_t|s) +w \nabla_{a_t} \mathcal{E}_t(s, a_t)$.
However, it poses issues to the performance stability of different $w$ during inference. 
Inspired by the experimental phenomenon that when the guidance scale is zero, the inference performance is more stable than that when the guidance scale is non-zero.
We propose to re-normalize the magnitude of $\nabla_{a_t}\log~\pi_t(a_t|s)$ by $\nabla_{a_t}\log~\pi_t(a_t|s) = \frac{\nabla_{a_t}\log~\pi_t(a_t|s)}{||\nabla_{a_t}\log~\pi_t(a_t|s)||}*||\nabla_{a_t}\log~\mu_t(a_t|s)||$.

\section{Experiments}

In the following sections, we report the details of environmental settings, evaluation metrics, and comparison results.

\begin{table*}[t!]
\centering
\small
\caption{Offline RL algorithms comparison on D4RL Gym-MuJoCo tasks, where we use \colorbox{myred}{red color} and \colorbox{myblue}{blue color} to show diffusion-based and non-diffusion-based baselines. Our method is shown with \colorbox{mycolor}{yellow color}.}
\label{Offline RL algorithms comparison on D4RL Gym-MuJoCo}
\resizebox{\textwidth}{!}{
\begin{tabular}{l | r r r | r r r | r r r | r | r}
\toprule
\specialrule{0em}{1.5pt}{1.5pt}
\toprule
Dataset & \multicolumn{3}{c|}{Med-Expert} & \multicolumn{3}{c|}{Medium} & \multicolumn{3}{c|}{Med-Replay} & \multirow{2}{*}{\makecell[r]{mean\\score}} & \multirow{2}{*}{\makecell[r]{total\\score}}\\
\cline{1-10}
\rule{0pt}{2.5ex} Env & HalfCheetah & Hopper & Walker2d & HalfCheetah & Hopper & Walker2d & HalfCheetah & Hopper & Walker2d &  &  \\
\midrule[1pt]
\rowcolor{myblue}
AWAC 
& 42.8 & 55.8 & 74.5 & 43.5 & 57.0 & 72.4 & 40.5 & 37.2 & 27.0 & 50.1 & 450.7\\
\rowcolor{myblue}
BC
& 55.2 & 52.5 & 107.5 & 42.6 & 52.9 & 75.3 & 36.6 & 18.1 & 26.0 & 51.9 & 466.7\\
\midrule
\rowcolor{myblue}
MOPO
& 63.3
& 23.7
& 44.6 
& 42.3
& 28.0 
& 17.8
& 53.1
& 67.5
& 39.0
& 42.1 & 379.3\\
\rowcolor{myblue}
MBOP
& 105.9 & 55.1 & 70.2 & 44.6 & 48.8 & 41.0 & 42.3 & 12.4 & 9.7 & 47.8 & 430.0 \\
\rowcolor{myblue}
MOReL
& 53.3 & 108.7 & 95.6  & 42.1 & 95.4 & 77.8 & 40.2 & 93.6 & 49.8 & 72.9 & 656.5 \\
\rowcolor{myblue}
TAP
& 91.8
& 105.5
& 107.4
& 45.0
& 63.4
& 64.9
& 40.8
& 87.3
& 66.8
& 74.8 & 672.9\\
\midrule
\rowcolor{myblue}
BEAR
& 51.7 & 4.0 & 26.0 & 38.6 & 47.6 & 33.2 & 36.2 & 10.8 & 25.3 & 30.4 & 273.4 \\
\rowcolor{myblue}
BCQ
& 64.7 & 100.9 & 57.5 & 40.7 & 54.5 & 53.1 & 38.2 & 33.1 & 15.0 & 50.9 & 457.7 \\
\rowcolor{myblue}
CQL
& 62.4 & 98.7 & 111.0 & 44.4 & 58.0 & 79.2 & 46.2 & 48.6 & 26.7 & 63.9 & 575.2\\
\midrule
\rowcolor{myblue}
TD3+BC
& 90.7 & 98.0 & 110.1 & 48.3 & 59.3 & 83.7 & 44.6 & 60.9 & 81.8 & 75.3 & 677.4\\
\rowcolor{myblue}
IQL
& 86.7 & 91.5 & 109.6 & 47.4 & 66.3 & 78.3 & 44.2 & 94.7 & 73.9 & 77.0 & 692.6 \\
\rowcolor{myblue}
PBRL
& 92.3
& 110.8
& 110.1
& 57.9
& 75.3
& 89.6
& 45.1
& 100.6
& 77.7
& 84.4 & 759.4\\
\midrule
\rowcolor{myblue}
DT
& 90.7 & 98.0 & 110.1 & 42.6 & 67.6 & 74.0 & 36.6 & 82.7 & 66.6 & 74.3 & 668.9\\
\rowcolor{myblue}
TT
& 95.0 & 110.0 & 101.9 & 46.9 & 61.1 & 79.0 & 41.9 & 91.5 & 82.6 & 78.9 & 709.9 \\
\rowcolor{myblue}
BooT
& 94.0 & 102.3 & 110.4 & 50.6 & 70.2 & 82.9 & 46.5 & 92.9 & 87.6 & 81.9 & 737.4 \\
\midrule
\rowcolor{myred}
SfBC
& 92.6
& 108.6
& 109.8
& 45.9
& 57.1
& 77.9
& 37.1
& 86.2
& 65.1
& 75.6 & 680.3\\
\rowcolor{myred}
D-QL@1
& 94.8 & 100.6 & 108.9 & 47.8 & 64.1 & 82.0 & 44.0 & 63.1 & 75.4 & 75.6 & 680.7\\
\rowcolor{myred}
Diffuser
& 88.9
& 103.3
& 106.9
& 42.8
& 74.3 
& 79.6
& 37.7
& 93.6
& 70.6
& 77.5 & 697.7\\
\rowcolor{myred}
DD
& 90.6
& 111.8
& 108.8
& 49.1
& 79.3
& 82.5
& 39.3 
& 100.0
& 75.0
& 81.8 & 736.4\\
\rowcolor{myred}
IDQL
& 95.9 & 108.6 & 112.7 & 51.0 & 65.4 & 82.5 & 45.9 & 92.1 & 85.1 & 82.1 & 739.2\\
\rowcolor{myred}
HDMI
& 92.1
& 113.5
& 107.9
& 48.0
& 76.4
& 79.9
& 44.9
& 99.6
& 80.7
& 82.6 & 743.0\\
\rowcolor{myred}
AdaptDiffuser
& 89.6
& 111.6
& 108.2
& 44.2
& 96.6
& 84.4
& 38.3
& 92.2
& 84.7 
& 83.3 & 749.8\\
\rowcolor{myred}
DiffuserLite
& 87.8
& 110.7
& 106.5
& 47.6
& 99.1
& 85.9
& 41.4
& 95.9
& 84.3
& 84.4 & {759.2}\\
\rowcolor{myred}
HD-DA
& 92.5
& 115.3
& 107.1
& 46.7
& 99.3
& 84.0
& 38.1
& 94.7
& 84.1
& 84.6 & 761.8 \\
\rowcolor{myred}
Consistency-AC
& 84.3
& 100.4
& 110.4
& 69.1
& 80.7
& 83.1
& 58.7
& 99.7
& 79.5
& 85.1 & 765.9\\
\rowcolor{myred}
TCD
& 92.7
& 112.6
& 111.3
& 47.2
& 99.4
& 82.1
& 40.6
& 97.2
& 88.0
& 85.7 & 771.0 \\
\rowcolor{myred}
D-QL
& 96.1 & 110.7 & 109.7 & 50.6 & 82.4 & 85.1 & 47.5 & 100.7 & 94.3 & 86.3 & 777.1\\
\rowcolor{myred}
QGPO
& 93.5
& 108.0
& 110.7
& 54.1
& 98.0
& 86.0
& 47.6
& 96.9
& 84.4
& 86.6 & 779.2\\
\midrule
\rowcolor{mycolor}
\ourmodel{}
& 94.4\tiny{$\pm$0.9}
& 111.5\tiny{$\pm$1.1}
& 109.3\tiny{$\pm$0.5}
& 49.6\tiny{$\pm$1.1}
& 100.2\tiny{$\pm$0.5} 
& 86.2\tiny{$\pm$1.1}
& 43.7\tiny{$\pm$1.3} 
& 101.0\tiny{$\pm$0.9}
& 90.8\tiny{$\pm$1.5}  
& 87.4
& 786.7
\\
\bottomrule
\specialrule{0em}{1.5pt}{1.5pt}
\bottomrule
\end{tabular}}
\vspace{-0.1cm}
\end{table*}

\subsection{Environments}

We select D4RL tasks~\cite{fu2020d4rl} as the test bed, which contains four types of benchmarks, Gym-MuJoCo, Pointmaze, Locomotion, and Adroit, with different dataset qualities.
Gym-MuJoCo tasks are composed of HalfCheetah, Hopper, and Walker2d with different difficulty settings datasets (e.g., medium, medium-replay, and medium-expert), where `medium-replay' and `medium-expert' denote the mixture level of behavior policies and `medium' represents uni-level behavior policy.
In Pointmaze, we select Maze2D with three difficulty settings (umaze, medium, and large) and two reward settings (sparse and dense) for evaluation.
Locomotion contains an ant robot control task with different maze sizes, which we distinguish using umaze, medium, and large.
These datasets include an abundant fraction of near-optimal episodes, making training challenging.
Adroit contains several sparse-reward, high-dimensional hand manipulation tasks where the datasets are collected under three types of situations (human, expert, and cloned).

\subsection{Metrics}

Considering the various reward structures of different environments, we select the normalized score as the comparison metric.
As an example, the normalized score $E_{norm}$ is calculated by $E_{norm}=\frac{E-E_{random}}{E_{expert}-E_{random}}*100$, where $E_{random}$ and $E_{expert}$ are the performance of random and expert policies, and $E$ is the evaluation performance.
In the ablation study, we select the radar chart to compare the holistic capacity of all methods, where the normalized performance is calculated by $E_{norm}=\frac{E_i}{\max(\{E_j|j\in{1,...,N}\})}$, where $E_i$ is the evaluation performance of method $i$, $i$ indicates the index of comparison methods, and $N$ is the total number of methods.

\begin{table*}[t!]
\centering
\vspace{-1em}
\small
\caption{D4RL Pointmaze (maze2d) and Locomotion (antmaze) comparison. We select 12 subtasks for evaluation, including different difficulties and reward settings. We use \colorbox{myred}{red color}, \colorbox{myblue}{blue color}, and \colorbox{mycolor}{yellow color} to show diffusion-based baselines, non-diffusion-based baselines, and our method.}
\label{mazed2d and antmaze evaluation}
\resizebox{\textwidth}{!}{
\begin{tabular}{l | r r | r r | r r | r r}
\toprule
\specialrule{0em}{1.5pt}{1.5pt}
\toprule
Environment & \multicolumn{2}{c|}{maze2d-umaze} & \multicolumn{2}{c|}{maze2d-medium} & \multicolumn{2}{c|}{maze2d-large} & \multirow{2}{*}{mean sparse score} & \multirow{2}{*}{mean dense score}\\
\cline{1-7}
\specialrule{0em}{1.5pt}{1.5pt}
Environment type & sparse & dense & sparse & dense & sparse & dense &  & \\
\midrule
\specialrule{0em}{1.5pt}{1.5pt}
 \rowcolor{myblue}
 DT      & 31.0  & - & 8.2 & - & 2.3 & - & 13.8 & - \\
 \rowcolor{myblue}
 BCQ     & 49.1 & - & 17.1 & - & 30.8 & - & 32.3 & - \\
 \rowcolor{myblue}
 QDT
 & 57.3  & - & 13.3  & - & 31.0  & - & 33.9 & - \\
 \rowcolor{myblue}
 IQL     & 42.1 & - & 34.9 & - & 61.7 & - & 46.2 & - \\
 \rowcolor{myblue}
 COMBO
 & 76.4 & - & 68.5 & - & 14.1 & - & 53.0 & -\\
 \rowcolor{myblue}
 TD3+BC  & 14.8 & - & 62.1 & - & 88.6 & - & 55.2 & - \\
 \rowcolor{myblue}
 BEAR    & 65.7 & - & 25.0 & - & 81.0 & - & 57.2 & - \\
  \rowcolor{myblue}
 BC      & 88.9 & 14.6 & 38.3 & 16.3 & 1.5  & 17.1 & 42.9 & 16.0\\
 \rowcolor{myblue}
 CQL     & 94.7 & 37.1 & 41.8 & 32.1 & 49.6 & 29.6 & 62.0 & 32.9\\
 \rowcolor{myblue}
 TT      & 68.7  & 46.6 & 34.9 & 52.7 & 27.6 & 56.6 & 43.7 & 52.0 \\
 \midrule
 \rowcolor{myred}
 SfBC    
 & 73.9
 & - 
 & 73.8
 & - 
 & 74.4
 & -
 & 74.0 & - \\
 \rowcolor{myred}
 SynthER
 & 99.1
 & -
 & 66.4
 & -
 & 143.3
 & - 
 & 102.9 & - \\
 \rowcolor{myred}
 Diffuser& 113.9 & - & 121.5 & - & 123.0 & - & 119.5 & - \\
 \rowcolor{myred}
 HDMI
 & 120.1
 & - 
 & 121.8
 & -
 & 128.6
 & - 
 & 123.5 & - \\
 \rowcolor{myred}
 HD-DA   & 72.8  & 45.5 & 42.1 & 54.7 & 80.7 & 45.7 & 65.2 & 48.6\\
 \rowcolor{myred}
 TCD    & 128.1 & 29.8 & 132.9 & 41.4 & 146.4 & 75.5 & 135.8 & 48.9\\
 \rowcolor{myred}
 DD      & 116.2 & 83.2 & 122.3 & 78.2 & 125.9 & 23.0 & 121.5 & 61.5 \\
 \midrule
 \rowcolor{mycolor}
 \ourmodel{}& 136.0 & 107.2 & 128.4 & 109.9 & 132.4 & 165.5 & 132.3 & 127.5 \\
\midrule[1pt]
Environment & \multicolumn{2}{c|}{antmaze-umaze} & \multicolumn{2}{c|}{antmaze-medium} & \multicolumn{2}{c|}{antmaze-large} & \multirow{2}{*}{mean score} & \multirow{2}{*}{total score}\\
\cline{1-7}
\specialrule{0em}{1.5pt}{1.5pt}
Environment type &  & diverse & play & diverse & play & diverse &  & \\
\midrule
\specialrule{0em}{1.5pt}{1.5pt}
\rowcolor{myblue}
 AWAC
 & 56.7 & 49.3 & 0.0 & 0.7 & 0.0 & 1.0 & 18.0 & 107.7\\
 \rowcolor{myblue}
 DT      & 59.2 & 53.0 & 0.0 & 0.0 & 0.0 & 0.0 & 18.7 & 112.2 \\
 \rowcolor{myblue}
 BC      & 65.0 & 55.0 & 0.0 & 0.0 & 0.0 & 0.0 & 20.0 & 120.0\\
 \rowcolor{myblue}
 BEAR
 & 73.0 & 61.0 & 0.0 & 8.0 & 0.0 & 0.0 & 23.7 & 142.0\\
 \rowcolor{myblue}
 BCQ     & 78.9 & 55.0 & 0.0 & 0.0 & 6.7 & 2.2 & 23.8 & 142.8\\
 \rowcolor{myblue}
 TD3+BC  & 78.6 & 71.4 & 10.6 & 3.0 & 0.2 & 0.0 & 27.3 & 163.8\\
 \rowcolor{myblue}
  CQL     & 74.0 & 84.0 & 61.2 & 53.7 & 15.8 & 14.9 & 50.6 & 303.6\\
 \rowcolor{myblue}
 IQL     & 87.5 & 62.2 & 71.2 & 70.0 & 39.6 & 47.5 & 63.0 & 378.0\\
 \midrule
  \rowcolor{myred}
 DD      & 73.1 & 49.2 & 0.0 & 24.6 & 0.0 & 7.5 & 25.7 & 154.4\\
 \rowcolor{myred}
 D-QL    
 & 93.4
 & 66.2
 & 76.6
 & 78.6
 & 46.4
 & 56.6
 & 69.6 & 417.8 \\
 \rowcolor{myred}
 IDQL    & 93.8 & 62.0 & 86.6 & 83.5 & 57.0 & 56.4 & 73.2 & 439.3 \\
 \rowcolor{myred}
 SfBC    
 & 92.0
 & 85.3
 & 81.3
 & 82.0
 & 59.3
 & 45.5
 & 74.2 & 445.4\\
 \rowcolor{myred}
 QGPO    
 & 96.4
 & 74.4
 & 83.6
 & 83.8
 & 66.6
 & 64.8
 & 78.3 & 469.6 \\
 \midrule
 \rowcolor{mycolor}
 \ourmodel{}
 & 100.0
 & 100.0
 & 76.7
 & 83.3
 & 56.7
 & 66.7
 & 80.6 & 483.4
 \\
\bottomrule
\specialrule{0em}{1.5pt}{1.5pt}
\bottomrule
\end{tabular}}
\vspace{-0.3cm}
\end{table*}

\subsection{Baselines}

For baselines, we compare our method with diffusion-based methods (DiffuserLite~\cite{dong2024diffuserlite}, HD-DA~\cite{chen2024simple}, IDQL~\cite{hansen2023idql}, AdaptDiffuser~\cite{liang2023adaptdiffuser}, Consistency-AC~\cite{ding2023consistency}, HDMI~\cite{li2023hierarchical}, TCD~\cite{hu2023instructed}, QGPO~\cite{lu2023contrastive}, SfBC~\cite{chen2022offline}, DD~\cite{ajay2022conditional}, Diffuser~\cite{janner2022planning}, D-QL~\cite{wang2022diffusion}, and D-QL@1, etc.) and non-diffusion-based methods, including traditional RL methods (AWAC~\cite{nair2020awac}, SAC~\cite{haarnoja2018soft}, and BC), model-based methods (TAP~\cite{jiang2022efficient}, MOReL~\cite{kidambi2020morel}, MOPO~\cite{yu2020mopo}, and MBOP~\cite{argenson2020model}), constraint-based methods (CQL~\cite{kumar2020conservative}, BCQ~\cite{fujimoto2019off}, and BEAR~\cite{kumar2019stabilizing}, etc.), uncertainty-based methods (PBRL~\cite{bai2022pessimistic}, TD3+BC~\cite{fujimoto2021minimalist}, and IQL~\cite{kostrikov2021offline}), and transformer-based methods (BooT~\cite{wang2022bootstrapped}, TT~\cite{janner2021offline}, and DT~\cite{chen2021decision}, etc.).
In summary, these baselines range from traditional offline RL methods to recent generative model-based methods, and we compared more than 30 competitive methods across over 30 subtasks.

\subsection{Results}

\begin{figure}[t!]
 \begin{center}
 \ifthenelse{\equal{\figureresolution}{low resolution}}
    {\includegraphics[angle=0,width=0.5\textwidth]{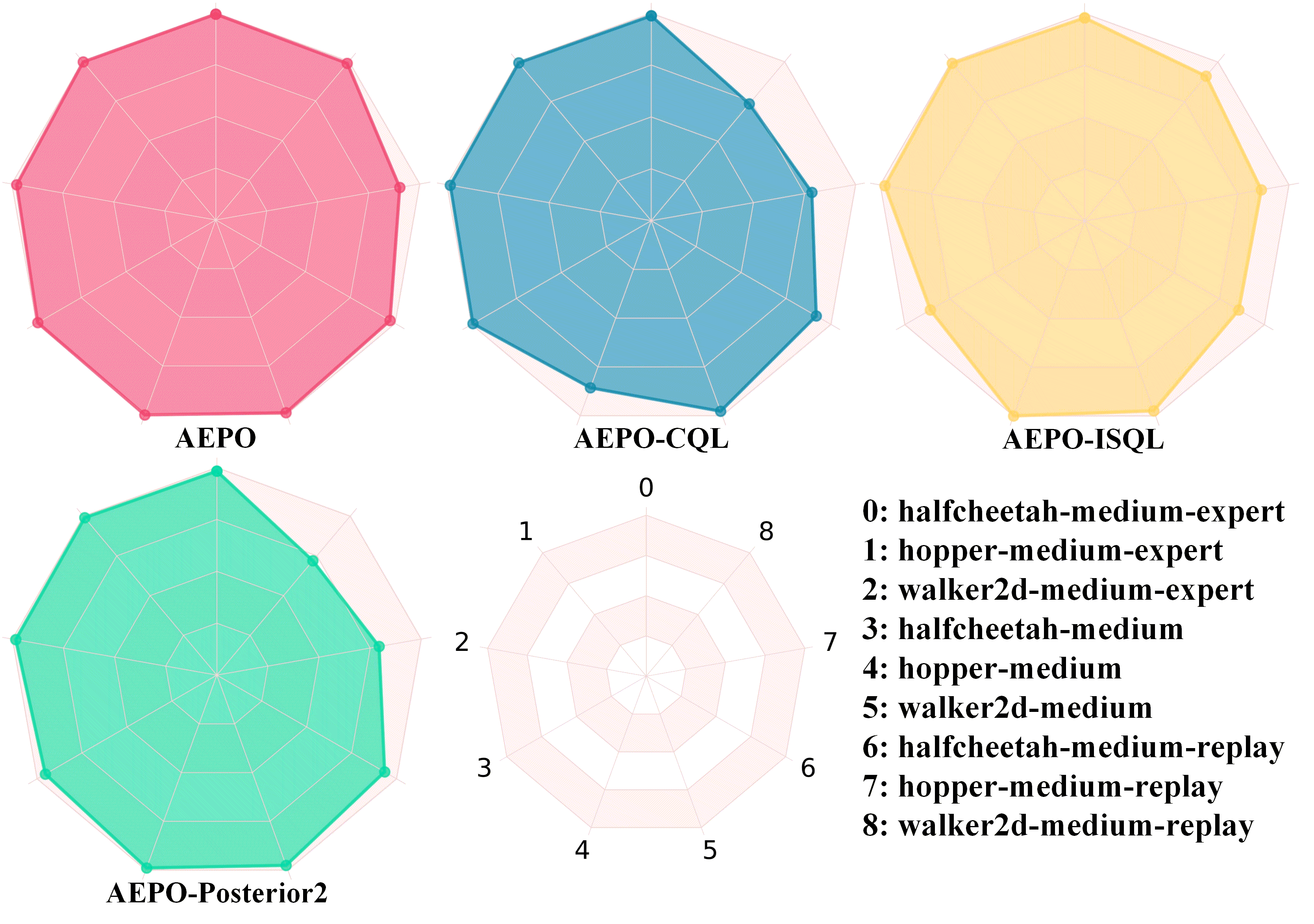}}
    {\includegraphics[angle=0,width=0.5\textwidth]{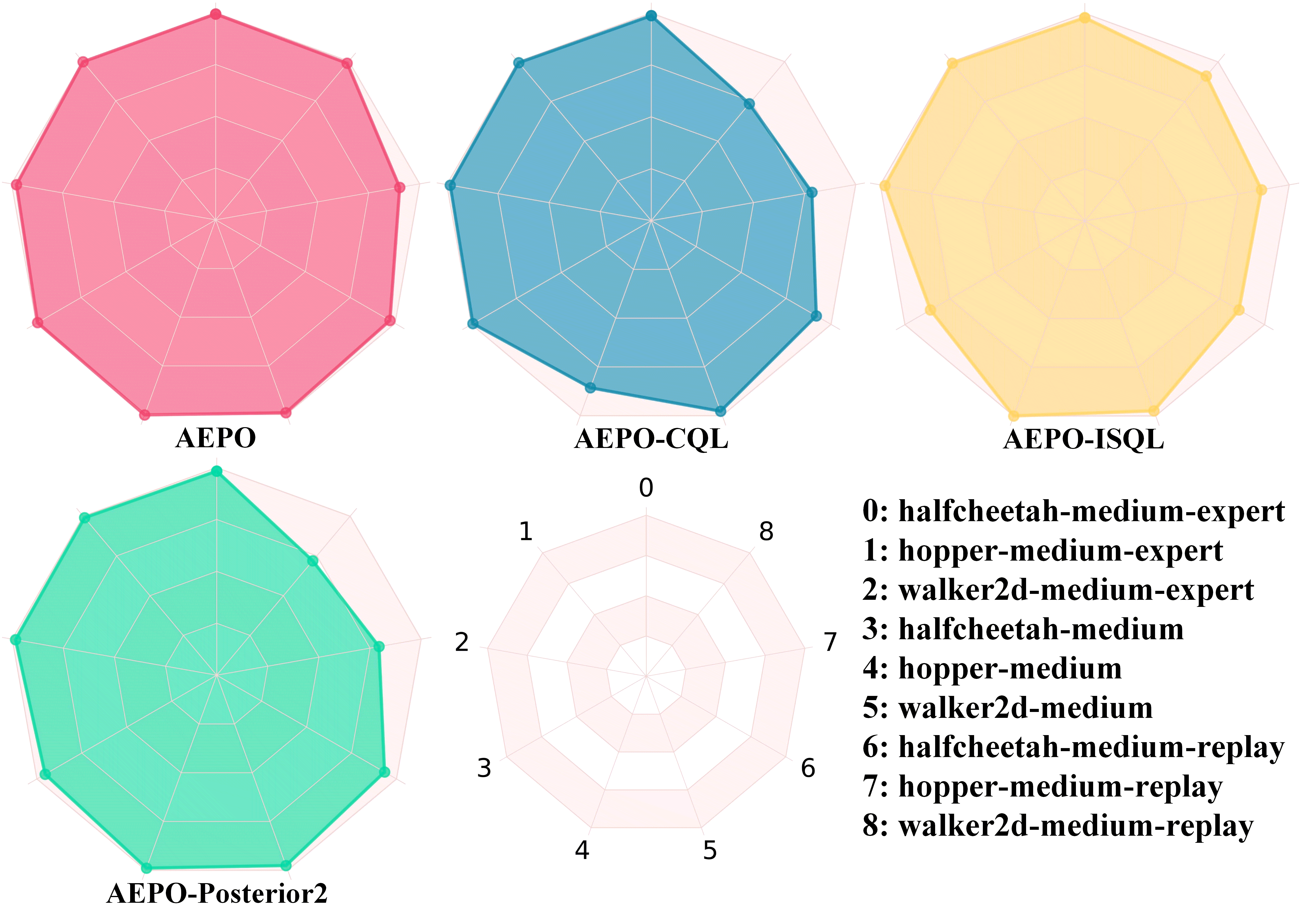}}
 \vspace{-0.3cm}
 \caption{Q function training and posterior approximation ablation of \ourmodel{} on D4RL Gym-MuJoCo tasks.}
 \label{D4RL_gym_mujoco_ablation}
 \end{center}
 \vspace{-0.4cm}
 \end{figure}

In Table~\ref{Offline RL algorithms comparison on D4RL Gym-MuJoCo}, we report the comparison results with dozens of diffusion-based and non-diffusion-based methods on the D4RL Gym-MuJoCo tasks.
Among all the algorithms on 9 tasks, the best performance of our method (\ourmodel{}) illustrates the effectiveness in decision-making scenarios.
Especially, in guided sampling, as a diffusion model, our method not only provides an approximate solution to guided sampling in theory but also surpasses most recent diffusion-based methods, such as QGPO, DiffuserLite, and D-QL, in abundant experiments.
In order to validate the effectiveness in sparse and dense reward settings, as well as hard tasks with the most sub-optimal trajectories, we conduct experiments on maze2d and antmaze tasks.
The results are reported in Table~\ref{mazed2d and antmaze evaluation}, where we can see that our method approaches or surpasses the SOTA algorithms.
Limited by the space, we postpone more experiments and discussion of D4RL Adroit in Table~\ref{Offline RL algorithms comparison on Adroit} of Appendix~\ref{Additional Experiments}.

Apart from Posterior 1 introduced in Section~\ref{Posterior Approximation}, we also investigate the performance of Posterior 2 on the D4RL Gym-MuJoCo tasks.
To show the performance differences obviously, we adopt the holistic performance and show them on the polygon shown in Figure~\ref{D4RL_gym_mujoco_ablation}, where each vertex represents a sub-task.
The fuller the polygon, the better the overall performance of the model.
From the figure, it can be observed that \ourmodel{} outperforms \ourmodel{}-Posterior 2. 
This may be attributed to the variance calculation.
As shown in Section~\ref{Posterior Approximation}, Posterior 2 is influenced by two factors—one being the intrinsic variance of the dataset and the other being the diffusion model. 
In contrast, Posterior 1 is influenced solely by the diffusion model. 
For the different Q function training strategies, we also conduct experiments on the Gym-MuJoCo tasks.
\ourmodel{}-CQL uses the contrastive q-learning to train the Q function, and \ourmodel{}-ISQL adopts in-support softmax q-learning to train the Q function.
The results in Figure~\ref{D4RL_gym_mujoco_ablation} illustrate that implicit Q-learning is an effective method to learn a better Q function in offline training.

\begin{figure}[t!]
 \begin{center}
 \ifthenelse{\equal{\figureresolution}{low resolution}}
    {\includegraphics[angle=0,width=0.48\textwidth]{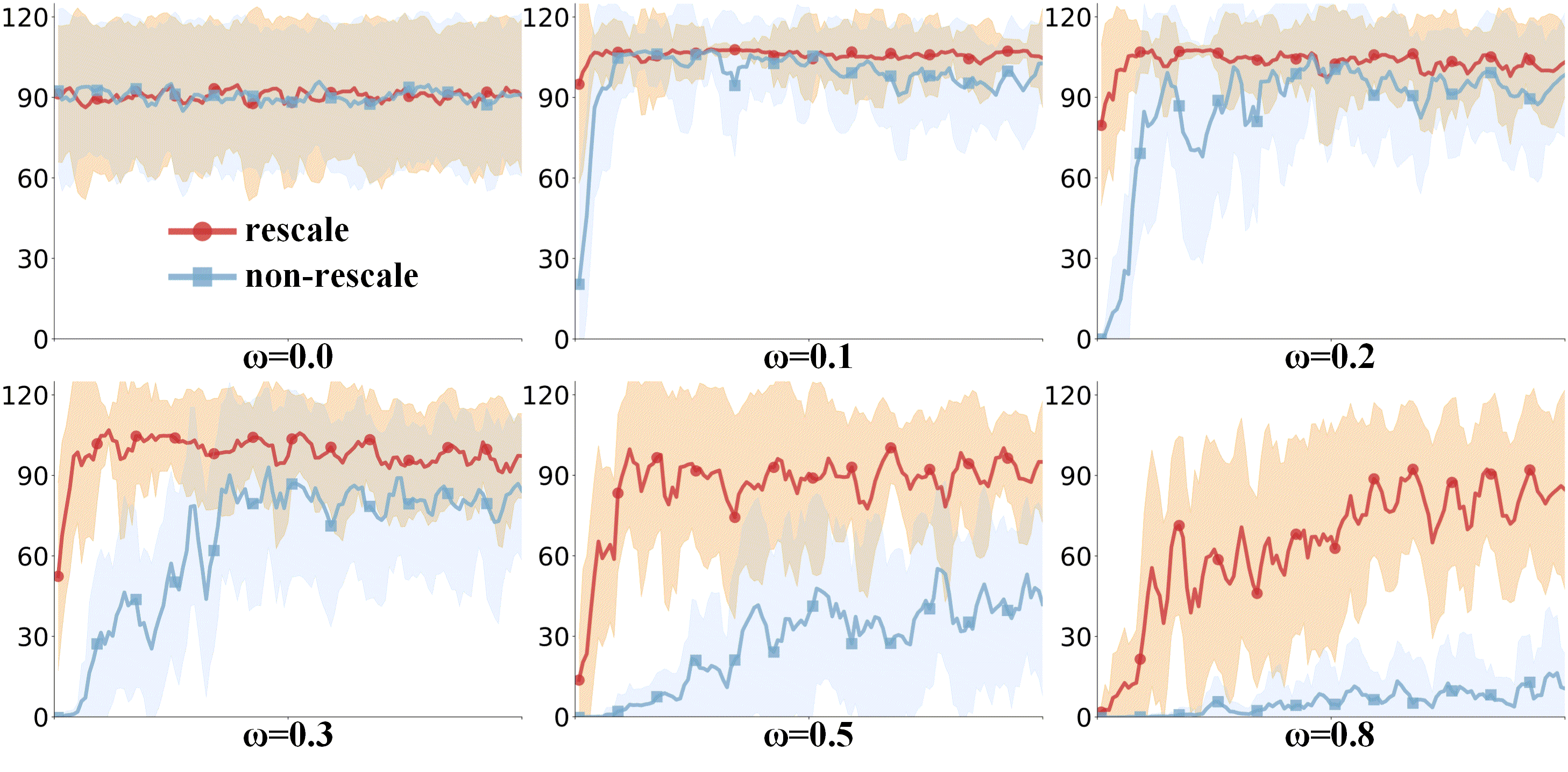}}
    {\includegraphics[angle=0,width=0.5\textwidth]{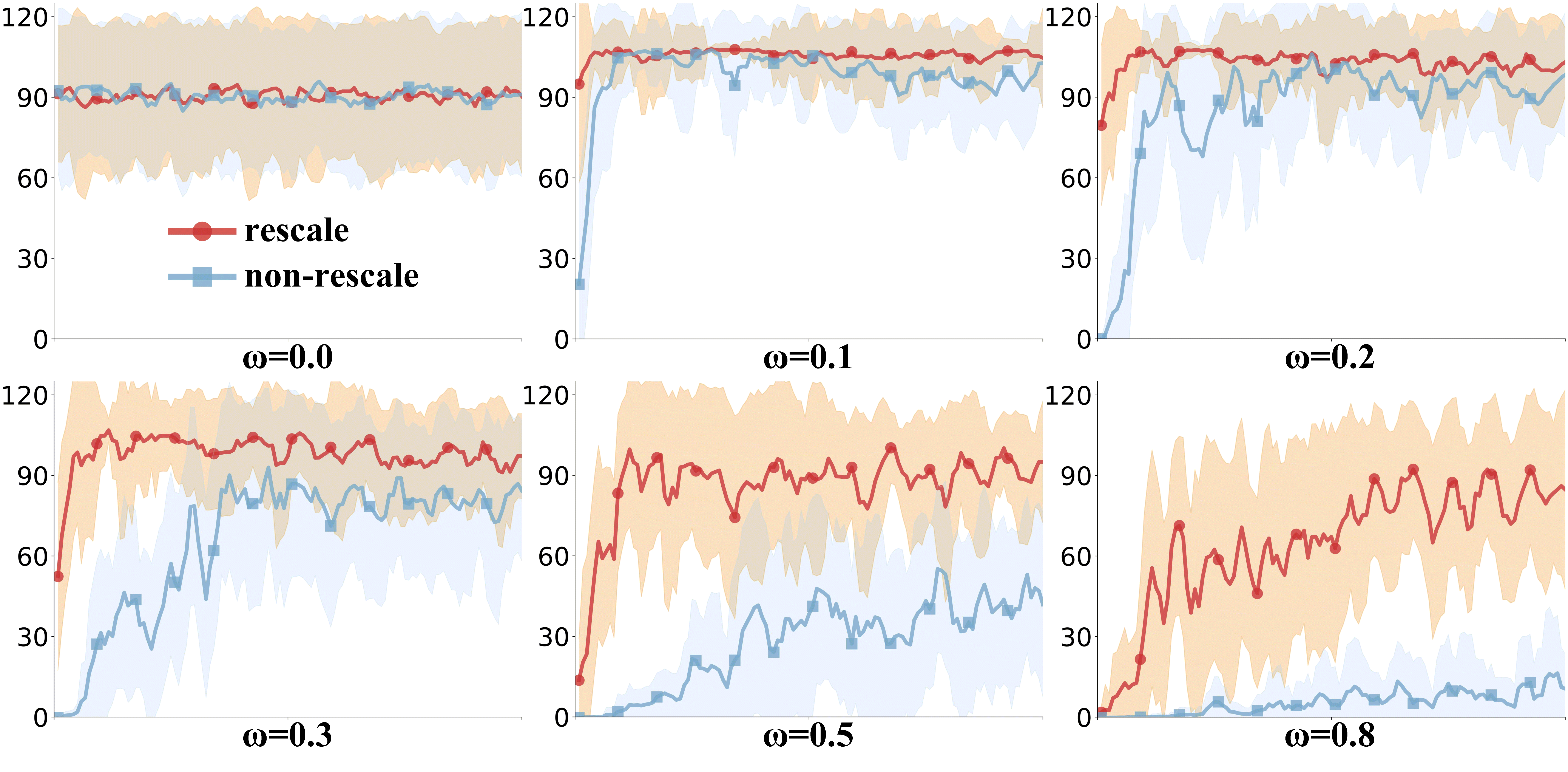}}
 \vspace{-0.3cm}
 \caption{Guidance rescale ablation of \ourmodel{} on Gym-MuJoCo walker2d-medium-expert task. The x-axis denotes training steps.}
\label{guidance_rescale_D4RL_gym_mujoco_ablation}
 \end{center}
 \vspace{-0.4cm}
 \end{figure}

As shown in Figure~\ref{guidance_rescale_D4RL_gym_mujoco_ablation}, guided sampling, such as $\omega = 0.1$, is important to reach higher returns compared with non-guided sampling ($\omega=0$).
With the increase of $\omega$, the performance of `non-rescale' decreases quickly, while `rescale' can still hold the performance. 
Obviously, guidance rescaling shows robustness to the guidance degree $\omega$, which makes it easier to find better performance in a wider range of $\omega$.

\section{Related Work}

\subsection{Offline RL}
Offline RL aims to learn a policy entirely from previously collected datasets, thus avoiding expensive and risky interactions with the environment, such as autonomous driving~\cite{levine2020offline, kumar2020conservative, kostrikov2021offline, liu2024selfbc, light2024dataset}. 
However, in practice, we may face the distribution shift issue, which means that the learned policy and the behavior policy are different, and the overestimation of out-of-distribution (OOD) actions will lead to a severe performance drop~\cite{kumar2020conservative, levine2020offline, fujimoto2021minimalist, ghosh2022offline, rezaeifar2022offline}. 
In order to solve these issues, previous studies can be roughly classified into several research lines. 
Policy regularization methods~\cite{beeson2024balancing, ma2024improving, yang2022behavior, fujimoto2021minimalist, kumar2019stabilizing} focus on applying constraints to the learned policy to prevent it from deviating far from the behavior policy. 
Critic penalty methods~\cite{kumar2020conservative, kostrikov2021offline, lyu2022mildly, mao2024supported} propose to train a conservative value function that assigns low expected return on unseen samples and high return on dataset samples, resulting in efficient OOD actions overestimation.
Uncertainty quantification methods~\cite{yu2020mopo, an2021uncertainty, bai2022pessimistic, wen2024towards} introduce uncertainty estimation to identify whether an action is OOD, thus enhancing the robustness on OOD actions.
Recently, generative models~\cite{he2023diffusion, li2023hierarchical, wang2024prioritized}, such as diffusion models, are proposed to augment the offline datasets with synthetic experiences or train a world model for planning.

\subsection{Generative Policy Optimization}
Recent advancements~\cite{janner2022planning} in diffusion RL methods have shown the diffusion models' powerful expression in modeling multimodal policies~\cite{chen2022offline, wang2022diffusion, pearce2023imitating}, learning heterogeneous behaviors~\cite{ajay2022conditional, hansen2023idql, li2024learning}, and generating fine-grained control~\cite{chi2023diffusion, dasari2024ingredients, zhang2024tedi}.
Diffusion-based RL algorithms cursorily contain two types of guided sampling to realize policy optimization~\cite{chen2023score}.
Guidance-based policy optimization~\cite{janner2022planning, lu2023contrastive, zhang2024revisiting} uses the value function to instruct the action generation process of diffusion models.
During the intermediate diffusion steps, the generated action vectors will incline to the region with high values according to the intermediate guidance.
Selected-based policy optimization~\cite{chen2022offline, hansen2023idql, mao2024diffusion} first generates a batch of candidate actions from diffusion behavior policy.
Then, it constructs a critic-weighted empirical action distribution and resamples the action from it for evaluation.

\section{Conclusion}
In this paper, we provide a theoretical analysis of the intermediate energy that matters in conditional decision generation with diffusion models.
We investigate the closed-form solution of intermediate guidance that has intractable log-expectation formulation and provide an effective approximation method under the most widely used Gaussian-based diffusion models.
Finally, we conduct sufficient experiments in 4 types, 30+ tasks by comparing with 30+ baselines to validate the effectiveness.


\nocite{langley00}

\bibliography{references}
\bibliographystyle{icml2025}

\newpage
\appendix
\onecolumn

\section{Pseudocode of \ourmodel{}}\label{Pseudocode}

\begin{algorithm}[h!]
\caption{Analytic Energy-guided Policy Optimization (\ourmodel{}).}
\label{algorithm}
\begin{algorithmic}[1]
\STATE \textbf{Input:} Dataset $D_\mu$, Max iterations $M$ of training, environmental time limit $\mathcal{T}$, generation steps $\mathcal{G}$, Noise prediction model $\epsilon_\theta$, intermediate energy model $\mathcal{E}_{\Theta}$, value function $V_\phi$, action-value function $Q_\psi$, target action-value function $Q_{\bar{\psi}}$
\STATE \textbf{Output:} Well-trained $\epsilon_\theta$, $\mathcal{E}_{\Theta}$, $V_\phi$, $Q_\psi$
\STATE \textcolor{gray}{// Training Process}
\FOR{$i=1$ $\textbf{to}$ $M$}
    \STATE \textcolor{gray}{// Train $\epsilon_\theta$ so that we can obtian $\nabla_{a_t}\log \mu_t(a_t|s)=-\epsilon_\theta(s, a_t, t)/\sigma_t$.}
    \STATE Train noise prediction model $\epsilon_\theta$ according to Equation~\eqref{general diffusion loss}
    \STATE \textcolor{gray}{// Train the Q function so that we can obtain $Q$ for any input $(s, a_0)$ and $Q^\prime$ with autograd.}
    \STATE Train $V_\phi$ and $Q_\psi$ with loss $\mathcal{L}_{V}$ and $\mathcal{L}_{Q}$ (Equation~\eqref{critic functions training loss})
    \STATE \textcolor{gray}{// For each sampled data from the dataset, we calculate the approximate intermediate energy value with the posterior distribution.}
    \STATE Calculate the target intermediate energy shown in Equation~\eqref{intermediate energy approximation with IGD}
    \STATE \textcolor{gray}{// Use finite data point to fit intermediate energy function $\mathcal{E}_\Theta(s,a_t,t)$ with neural network.}
    \STATE Update the parameter $\Theta$ according to Equation~\eqref{intermediate energy loss}
\ENDFOR
\STATE // \textbf{Evaluation Process}
\FOR{$i=1$ $\textbf{to}$ $\mathcal{T}$}
    \STATE Receive state $s$ from the environment
    \STATE Sample $a_T$ from $\mathcal{N}(0,\bm{I})$
    \FOR{$t=T$ $\textbf{to}$ 0}
        \STATE Calculate $\nabla_{a_t}\log\mu_t(a_t|s)$ with Equation~\eqref{unguided intermediate guidance}
        \STATE Obtain $\nabla_{a_t}\mathcal{E}_\Theta(s,a_t,t)$ by performing gradient on $a_t$
        \STATE \textcolor{gray}{// $\nabla_{a_t}\log \pi_t(a_t|s) = \nabla_{a_t} \log~\mu_t(a_t|s) + \nabla_{a_t} \mathcal{E}_\Theta(s, a_t,t)$}
        \STATE Obtain score function $\nabla_{a_t}\log\pi_t(a_t|s)$ according to Equation~\eqref{intermediate energy_guidance in RL}
        \STATE \textcolor{gray}{// The implementation follows DPM-solver to realize lower generation steps and reduce time consumption.}
        \STATE Perform action denoising with Equation~\eqref{neural ODE}
    \ENDFOR
    \STATE Interact with the environment with generated action $a_0$
\ENDFOR
\end{algorithmic}
\end{algorithm}

The training and evaluation of \ourmodel{} are shown in Algorithm~\ref{algorithm}.
In lines 3-13, we sample data from the dataset and train the noise prediction model $\epsilon_\theta$ that is used to obtain $\nabla_{a_t}\log \mu_t(a_t|s)=-\epsilon_\theta(s, a_t, t)/\sigma_t$, the intermediate energy $\mathcal{E}_\Theta(s,a_t,t)$ that can be used to approximate the intermediate guidance $\nabla_{a_t}\mathcal{E}_t(s,a_t)$, and the Q function $Q_{\psi}$ that is used to calculate $\mathcal{E}_\Theta(s,a_t,t)$.
During the evaluation (lines 14-27), for each state received from the environment, we perform generation with Equation~\eqref{neural ODE}, where we use DPM-solver~\cite{lu2022dpm} as implementation.
After several generation steps, we obtain the generated action $a_0$ to interact with the environment.

\section{Additional Experiments}\label{Additional Experiments}

\subsection{Additional Experiments on D4RL Adroit}

D4RL Adroit is a hand-like robotic manipulation benchmark, which contains several sparse rewards and high-dimensional robotic manipulation tasks where the datasets are collected under three types of situations (-human, -expert, and -cloned)~\cite{rajeswaran2017learning}.
For example, the Pen is a scenario where the agent needs to get rewards by twirling a pen. The Relocate scenario requires the agent to pick up a ball and move it to a specific location.
The experiments of D4RL Adroit are shown in Table~\ref{Offline RL algorithms comparison on Adroit}, where we compare our method with non-diffusion-based methods highlighted with blue color and diffusion-based methods highlighted with red color.
Our method is highlighted in yellow color.
The results show that our method surpasses or matches the best performance in 11 of the total 12 sub-tasks, which illustrates strong competitiveness of our method.

\begin{table*}[h]
\centering
\small
\caption{Offline RL algorithms comparison on Adroit. We select 4 tasks for evaluation, where each task contains 3 types of difficulty settings. We use \colorbox{myred}{red color}, \colorbox{myblue}{blue color}, and \colorbox{mycolor}{yellow color} to show diffusion-based baselines, non-diffusion-based baselines, and our method.}
\label{Offline RL algorithms comparison on Adroit}
\resizebox{\textwidth}{!}{
\begin{tabular}{l | r r r | r r r | r r r | r r r | r r}
\toprule
\specialrule{0em}{1.5pt}{1.5pt}
\toprule
Task & \multicolumn{3}{c|}{pen} & \multicolumn{3}{c|}{hammer} & \multicolumn{3}{c|}{door} & \multicolumn{3}{c|}{relocate} & \makecell[r]{mean\\score} & \makecell[r]{total\\score}\\
\midrule
\rule{0pt}{2.5ex} Dataset & human & expert & cloned & human & expert & cloned & human & expert & cloned & human & expert & cloned &  & \\
\midrule[1pt]
\rowcolor{myblue}
BC 
& 7.5 & 69.7 & 6.6 & - & - & - & - & - & - & 0.1 & 57.1 & 0.1 & - & - \\
\rowcolor{myblue}
BEAR 
& -1.0 & - & 26.5 & - & - & - & - & - & - & - & - & - & - & - \\
\rowcolor{myblue}
BCQ 
& 68.9 & - & 44.0 & - & - & - & - & - & - & - & - & - & - & - \\
\rowcolor{myblue}
TT 
& 36.4 & 72.0 & 11.4 & 0.8 & 15.5 & 0.5 & 0.1 & 94.1 & -0.1 & 0.0 & 10.3 & -0.1 & 20.1 & 240.9 \\
\rowcolor{myblue}
CQL 
& 37.5 & 107.0 & 39.2 & 4.4 & 86.7 & 2.1 & 9.9 & 101.5 & 0.4 & 0.2 & 95.0 & -0.1 & 40.3 & 483.8 \\
\rowcolor{myblue}
TAP 
& 76.5
& 127.4
& 57.4
& 1.4
& 127.6
& 1.2
& 8.8
& 104.8
& 11.7
& 0.2
& 105.8
& -0.2
& 51.9 & 622.6 \\
\midrule
\rowcolor{myred}
IQL & 71.5 & - & 37.3 & 1.4 & - & 2.1 & 4.3 & - & 1.6 & 0.1 & - & -0.2 & - & - \\
\rowcolor{myred}
TCD 
& 49.9 & 35.6 & 73.3 & - & - & - & - & - & - & 0.4 & 59.6 & 0.2 & - & - \\
\rowcolor{myred}
HD-DA & -2.6 & 107.9 & -2.7 & - & - & - & - & - & - & 0.0 & -0.1 & -0.2 & - & - \\
\rowcolor{myred}
DiffuserLite & 33.2 & 20.7 & 2.1 & - & - & - & - & - & - & 0.1 & 0.1 & -0.2 & - & - \\
\rowcolor{myred}
DD 
& 64.1
& 107.6
& 47.7
& 1.0
& 106.7
& 0.9
& 6.9
& 87.0
& 9.0
& 0.2
& 87.5
& -0.2
& 43.2 & 518.4 \\
\rowcolor{myred}
HDMI 
& 66.2
& 109.5
& 48.3
& 1.2
& 111.8
& 1.0
& 7.1
& 85.9
& 9.3
& 0.1
& 91.3
& -0.1
& 44.3 & 531.6 \\
\rowcolor{myred}
D-QL@1 
& 66.0
& 112.6
& 49.3
& 1.3
& 114.8
& 1.1
& 8.0
& 93.7
& 10.6
& 0.2
& 95.2
& -0.2
& 46.1 & 552.6 \\
\rowcolor{myred}
QGPO 
& 73.9
& 119.1
& 54.2
& 1.4
& 123.2
& 1.1
& 8.5
& 98.8
& 11.2
& 0.2
& 102.5
& -0.2
& 49.5 & 593.9 \\
\rowcolor{myred}
LD 
& 79.0
& 131.2
& 60.7
& 4.6
& 132.5
& 4.2
& 9.8
& 111.9
& 12.0
& 0.2
& 109.5
& -0.1
& 54.6 & 655.5 \\
\midrule
\rowcolor{mycolor}
\ourmodel{} 
& 76.7
& 147.0
& 69.3
& 10.3
& 129.7
& 6.4
& 9.0
& 106.5
& 3.9
& 0.8
& 107.0
& 0.6
& 55.6 & 667.5 \\
\bottomrule
\specialrule{0em}{1.5pt}{1.5pt}
\bottomrule
\end{tabular}}
\vspace{-0.1cm}
\end{table*}

\subsection{Additional Rescaling Ablation}

\begin{figure}[t!]
 \begin{center}
 \ifthenelse{\equal{\figureresolution}{low resolution}}
    {\includegraphics[angle=0,width=0.9\textwidth]{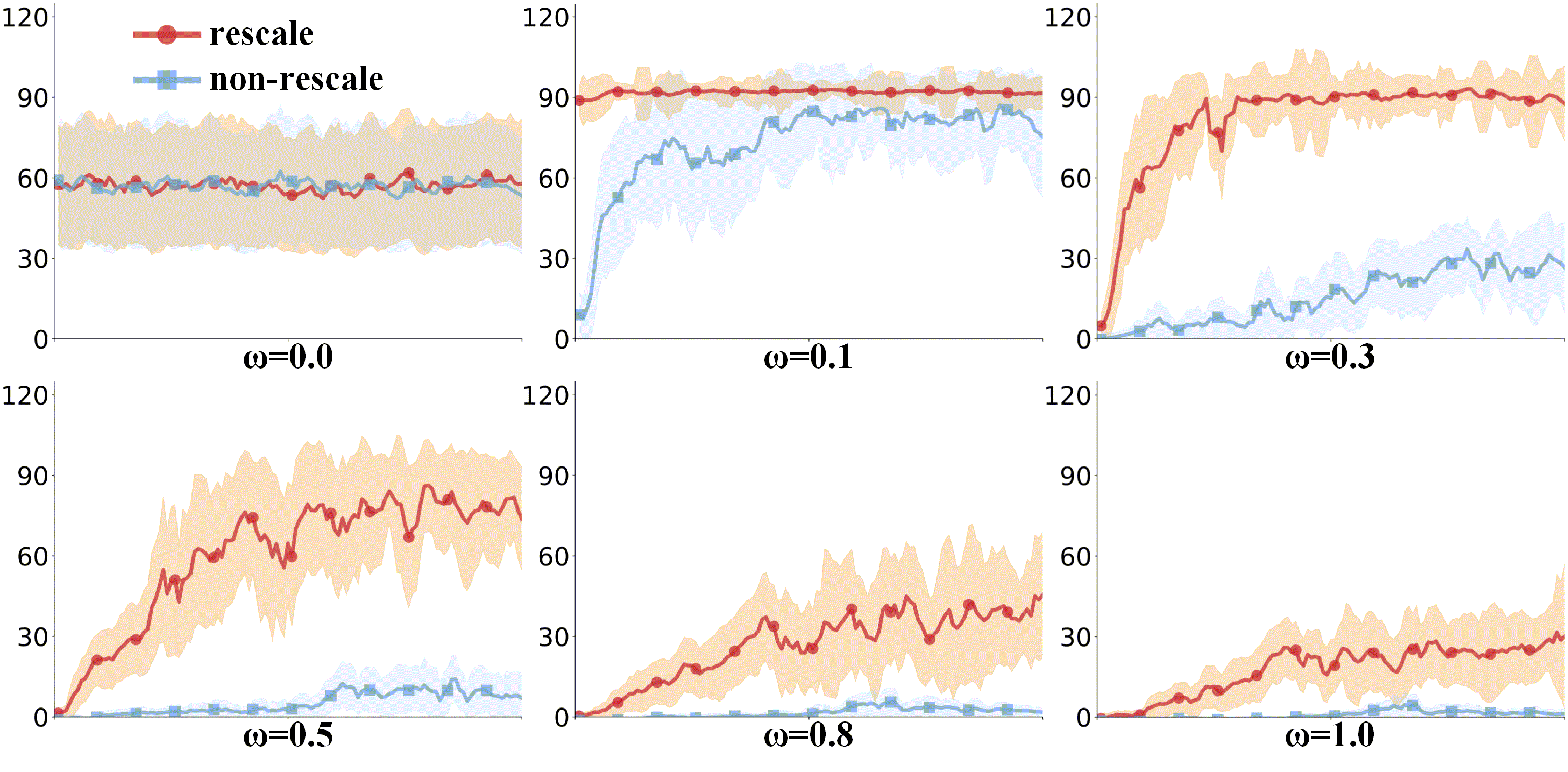}}
    {\includegraphics[angle=0,width=0.9\textwidth]{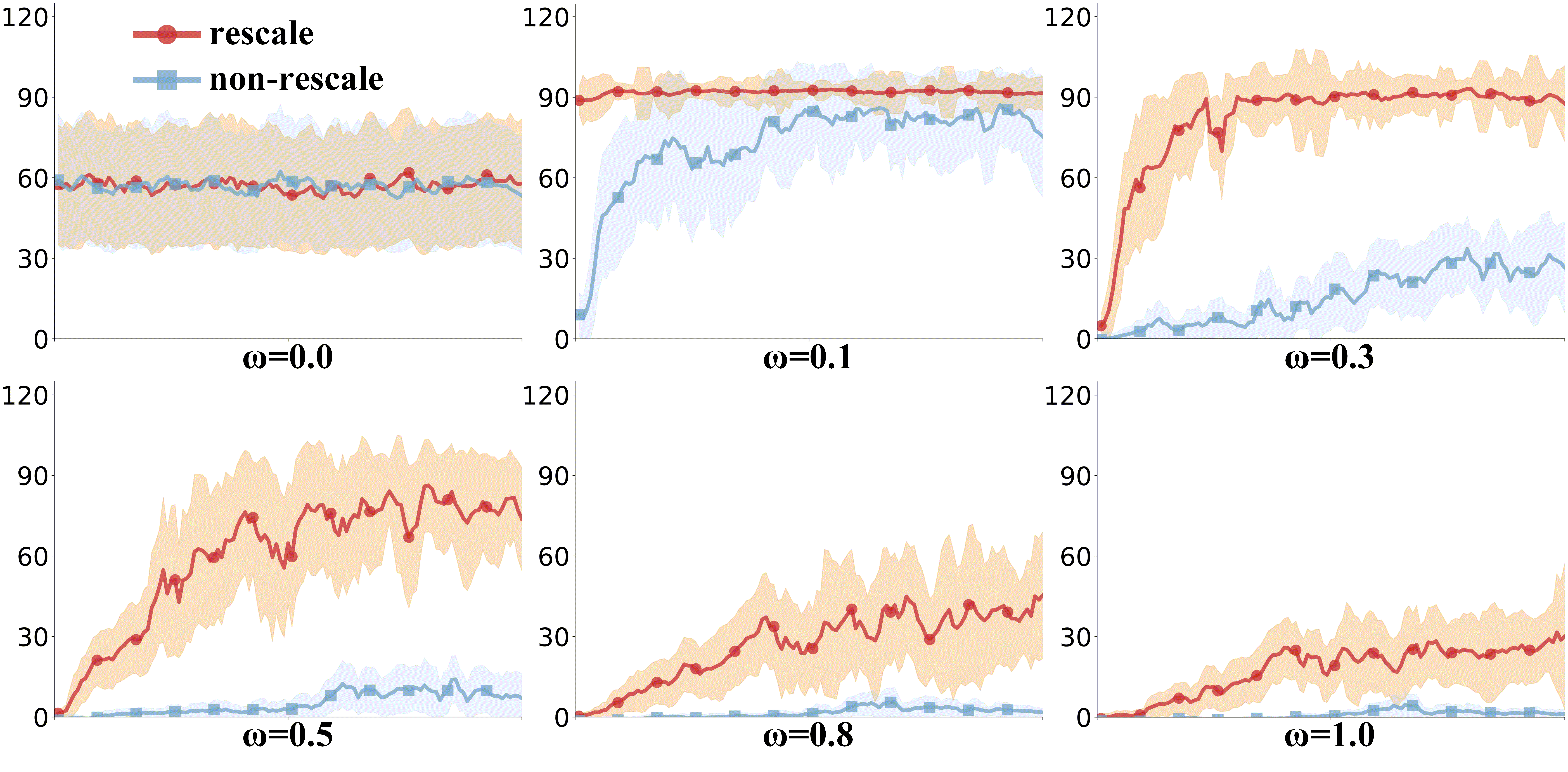}}
 \caption{Guidance rescale ablation of \ourmodel{} on D4RL Gym-MuJoCo halfcheetah-medium-expert task. The y-axis and x-axis denote the normalized score and training steps, respectively.}
 \label{rescale ablation on halfcheetah-me}
 \end{center}
 \end{figure}

\begin{figure}[t!]
 \begin{center}
 \ifthenelse{\equal{\figureresolution}{low resolution}}
    {\includegraphics[angle=0,width=0.9\textwidth]{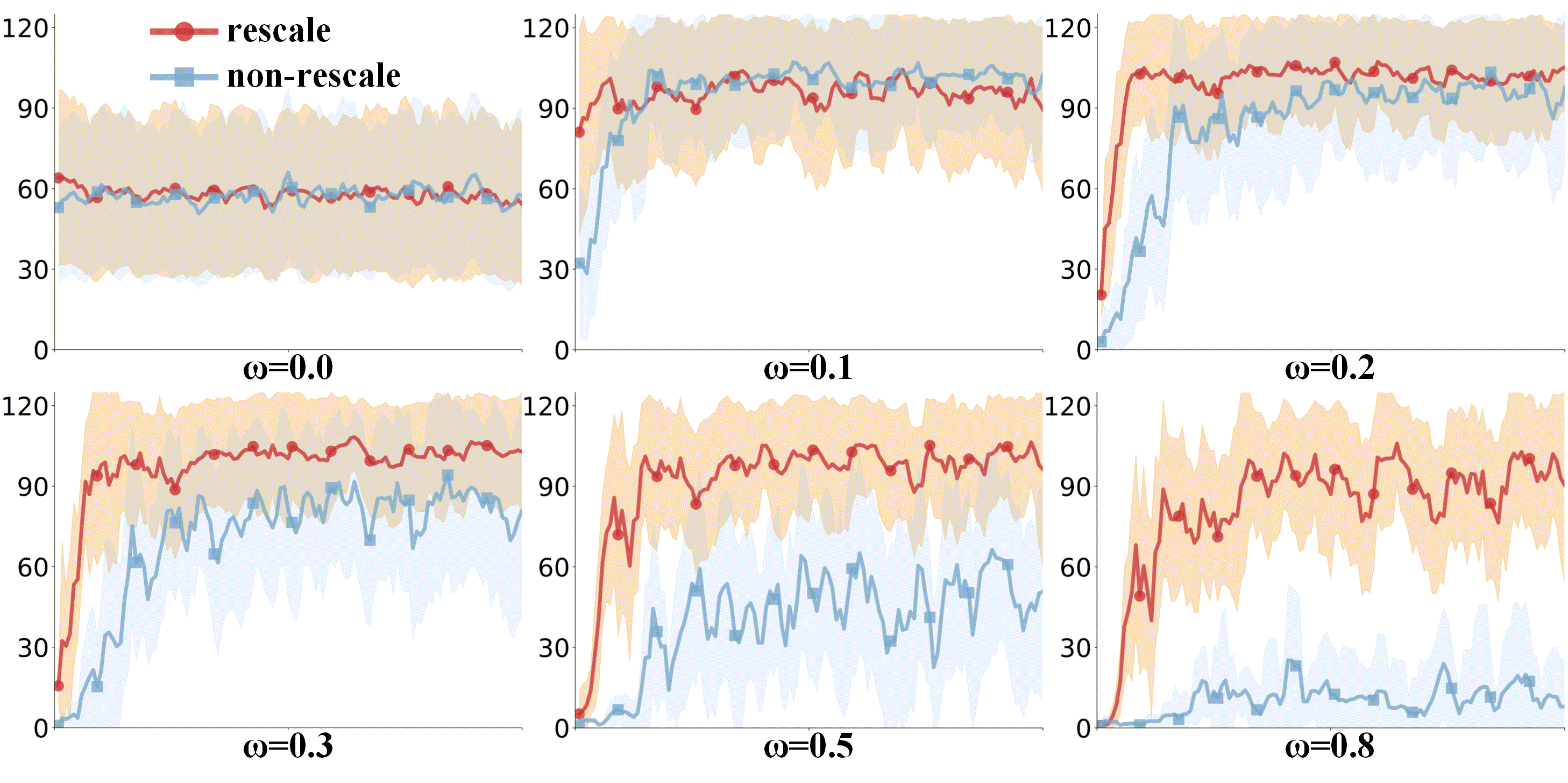}}
    {\includegraphics[angle=0,width=0.9\textwidth]{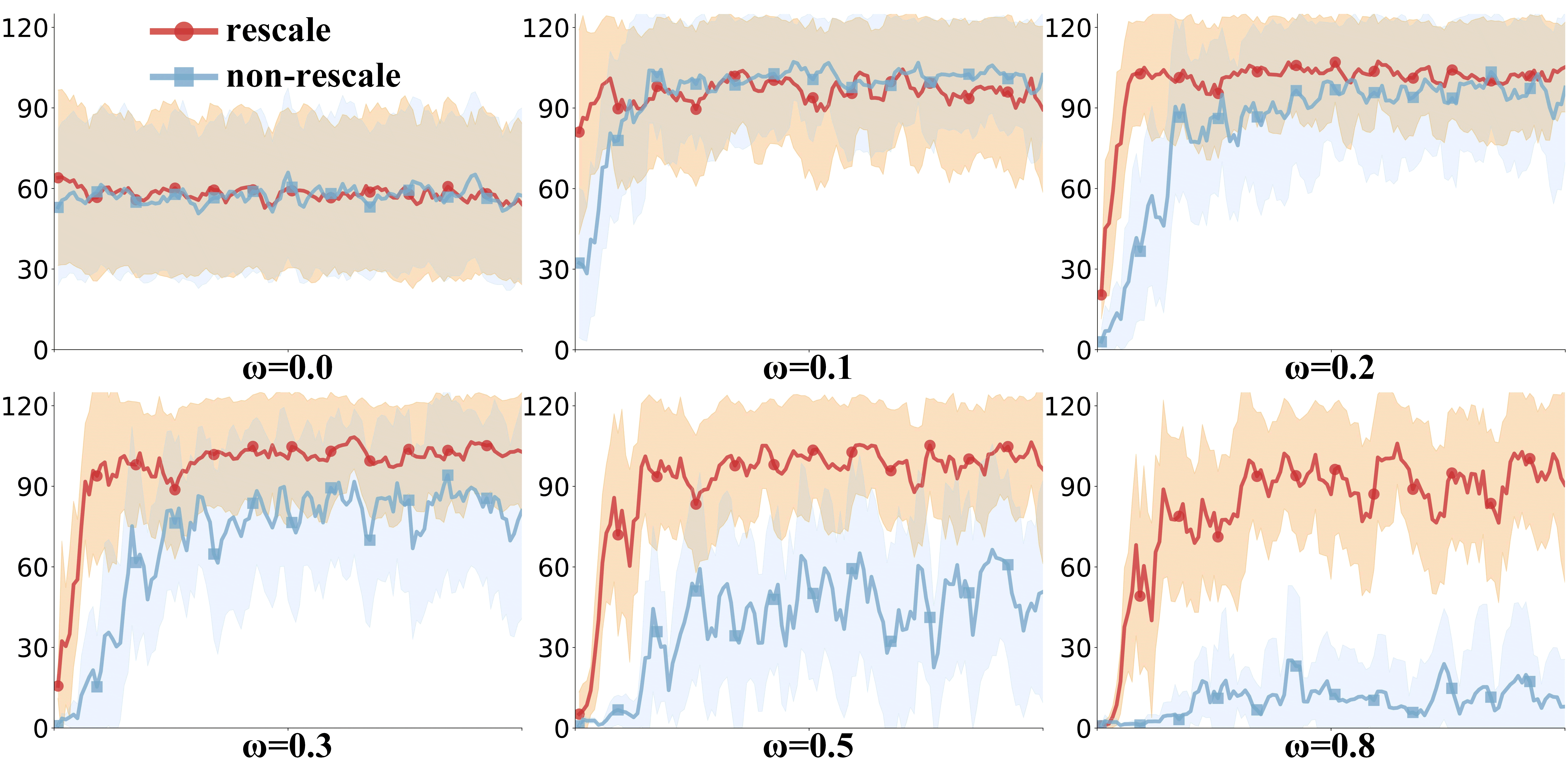}}
 \caption{Guidance rescale ablation of \ourmodel{} on D4RL Gym-MuJoCo hopper-medium-expert task. The y-axis and x-axis denote the normalized score and training steps, respectively.}
 \label{rescale ablation on hopper-me}
 \end{center}
 \end{figure}

\begin{figure}[t!]
 \begin{center}
 \ifthenelse{\equal{\figureresolution}{low resolution}}
    {\includegraphics[angle=0,width=0.9\textwidth]{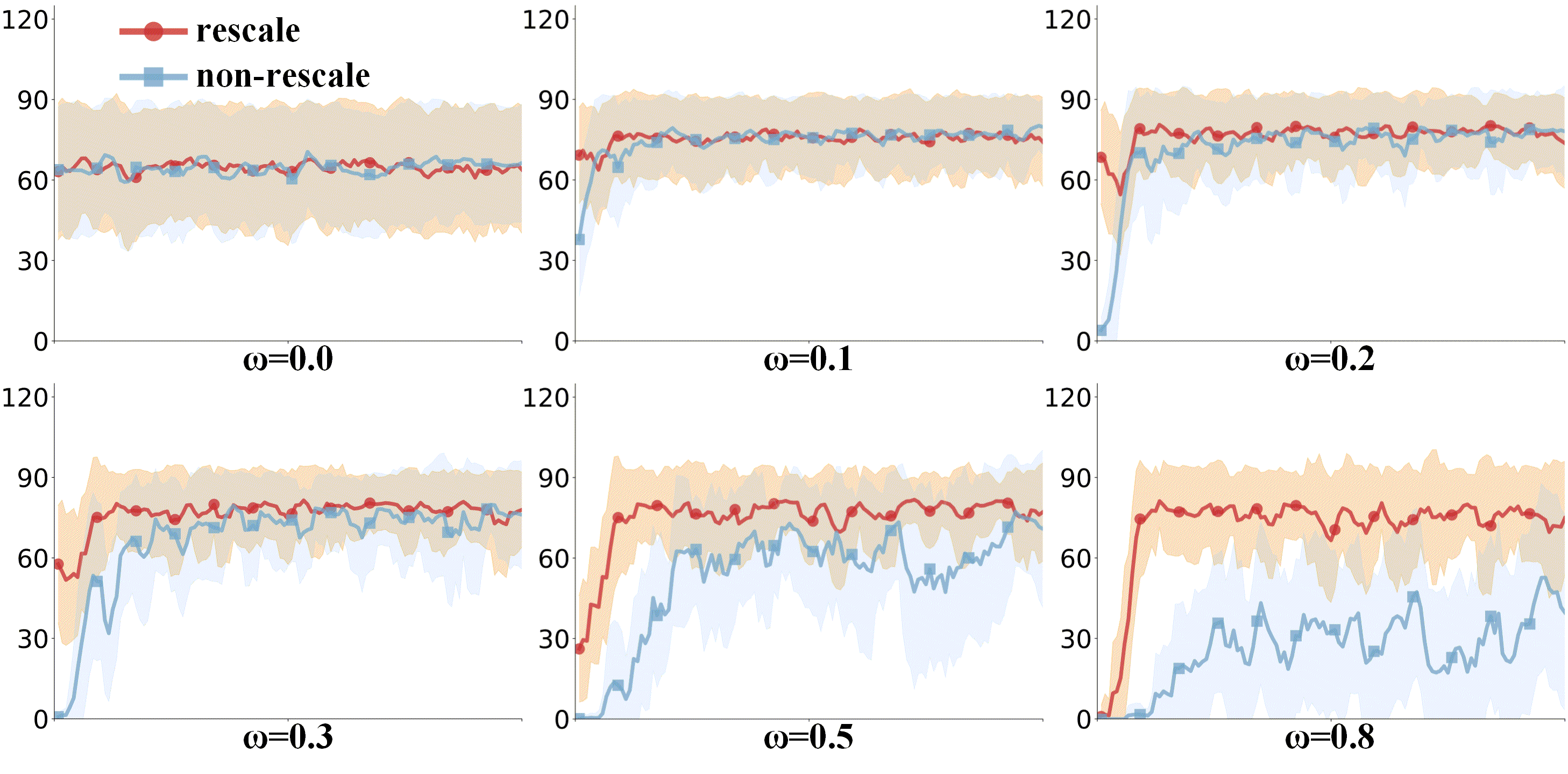}}
    {\includegraphics[angle=0,width=0.9\textwidth]{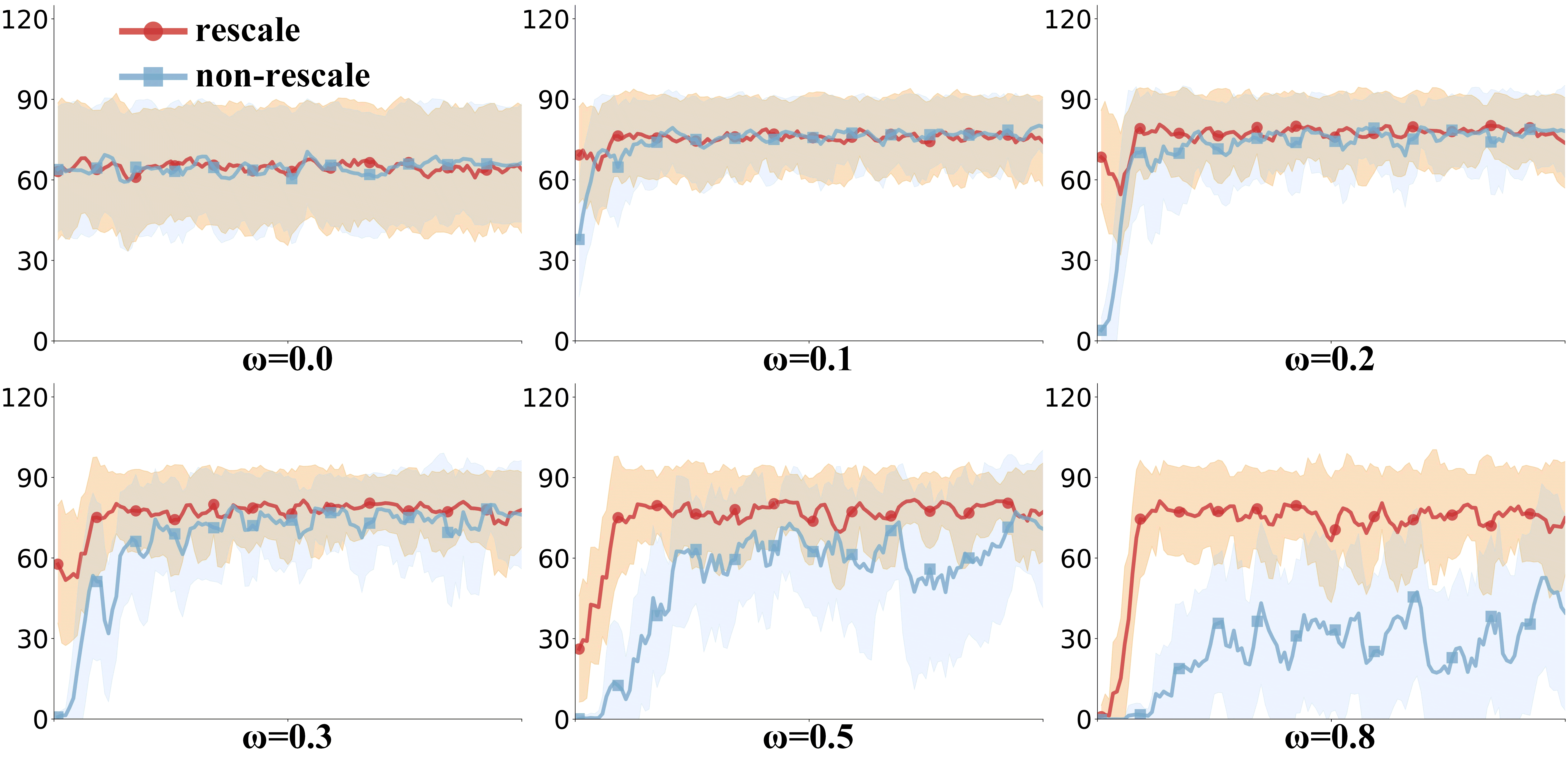}}
 \caption{Guidance rescale ablation of \ourmodel{} on D4RL Gym-MuJoCo walker2d-medium task. The y-axis and x-axis denote the normalized score and training steps, respectively.}
 \label{rescale ablation on walker2d-m}
 \end{center}
 \end{figure}

Apart from the guidance rescaling ablation on walker2d-medium-expert task, we also conduct the experiments on halfcheetach-medium-expert, hopper-medium-expert, and walker2d-medium tasks, where the results are shown in Figure~\ref{rescale ablation on halfcheetah-me}, Figure~\ref{rescale ablation on hopper-me}, and Figure~\ref{rescale ablation on walker2d-m}, respectively.
From the results, we can see that guidance is important to reach a better performance by comparing $\omega=0.1$ and $\omega=0$, where $\omega=0$ means no intermediate guidance.
`rescale' means we apply the guidance rescaling strategy, and `non-rescale' means we do not use guidance rescaling.
With the increase of $\omega$, the performance of `non-rescale' decreases quickly, while `rescale' can still hold the performance, which illustrates that guidance rescaling makes the model robust to the guidance degree $\omega$, thus leading to better performance in a wider range of $\omega$.

\subsection{Parameter Sensitivity}

\begin{figure}[t!]
 \begin{center}
 \ifthenelse{\equal{\figureresolution}{low resolution}}
    {\includegraphics[angle=0,width=0.9\textwidth]{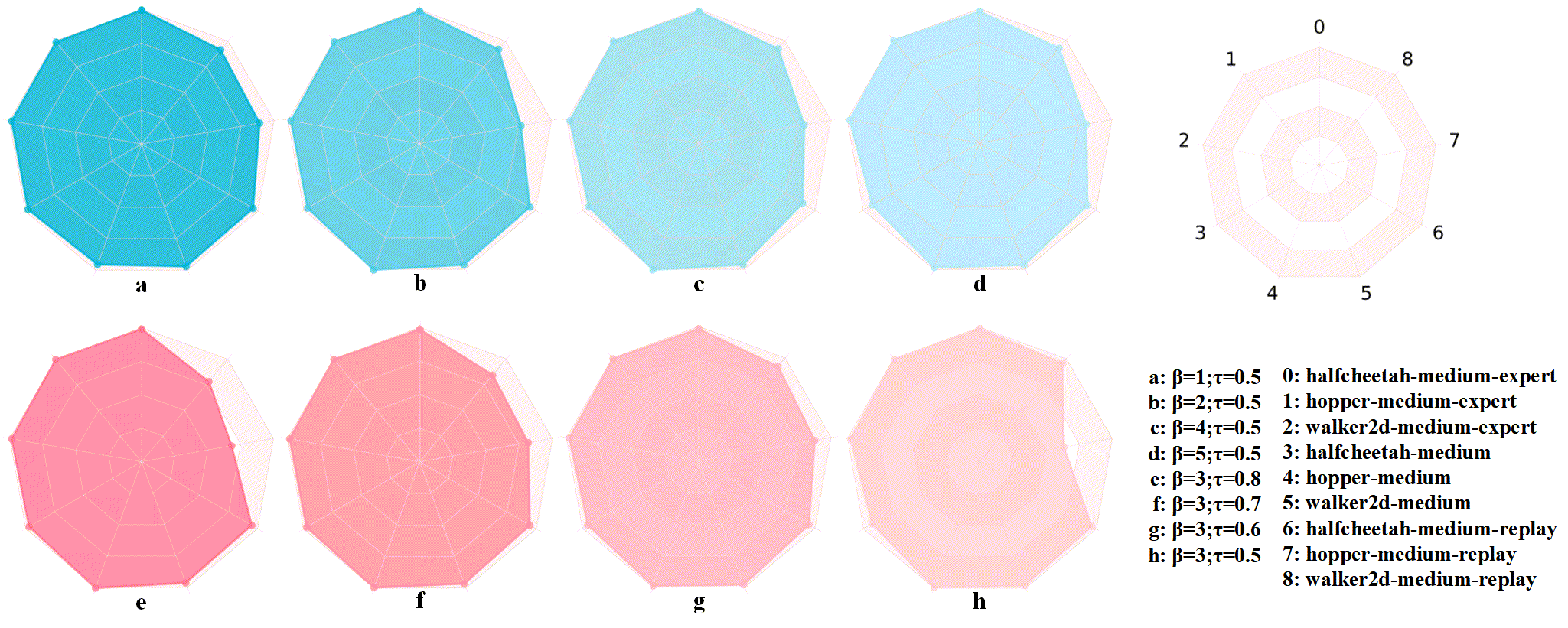}}
    {\includegraphics[angle=0,width=0.9\textwidth]{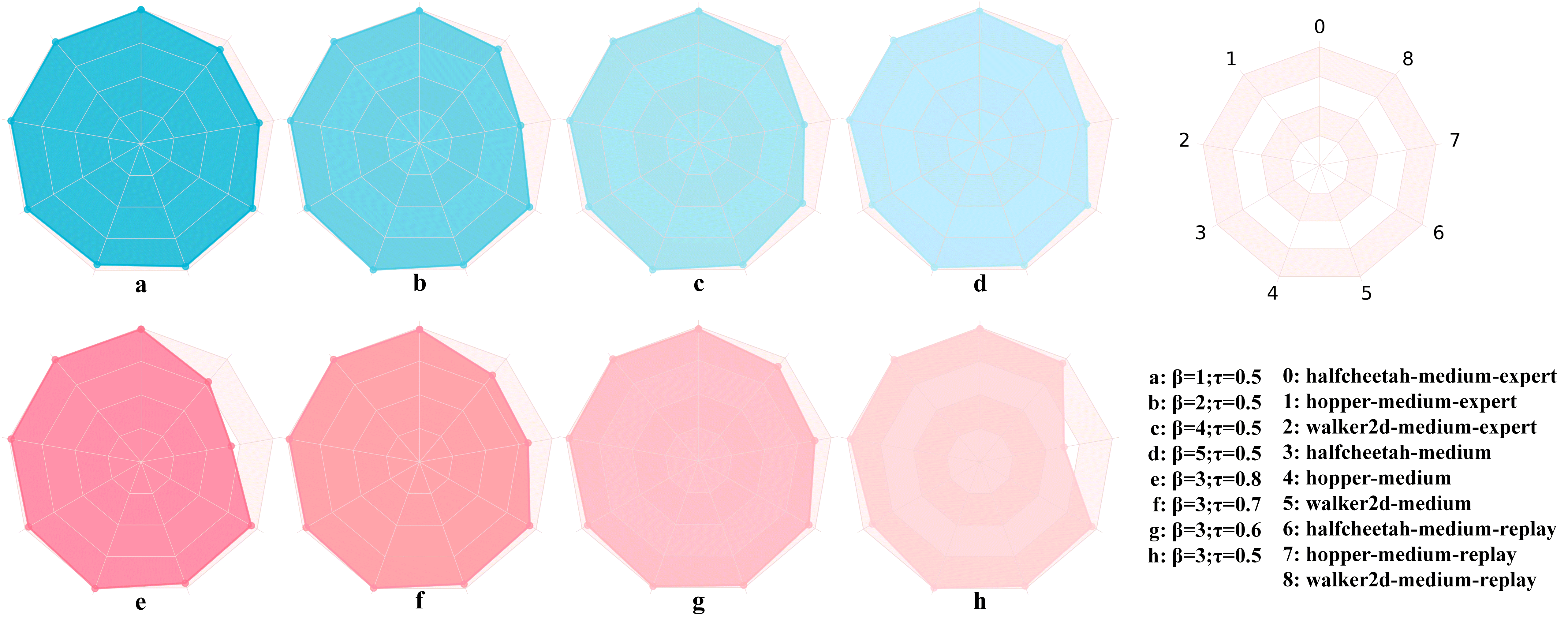}}
 \caption{Parameter sensitivity of  D4RL Gym-MuJoCo tasks.}
 \label{para sens on gym-mujoco}
 \end{center}
 \end{figure}

We investigate the influence of hyperparameters of $\beta$ and $\tau$ that are used in loss $\mathcal{L}_{IE}$ (Equation~\eqref{intermediate energy loss}) and $\mathcal{L}_{Q}$ (Equation~\eqref{critic functions training loss}).
From the results shown in Figure~\ref{para sens on gym-mujoco}, we can see that our method shows slight sensitivity to $\beta$ value.
For the hyperparameter $\tau$, we find that a certain value is friendly for achieving good performance, such as $\tau=0.7$ (Figure~\ref{para sens on gym-mujoco} (g)).

\begin{table}[t!]
\centering
\caption{The hyperparameters of \ourmodel{}.}
\label{our method hyper}
\begin{tabular}{l l}
\toprule
Hyperparameter & Value\\
\midrule
network backbone & MLP\\
action value function ($Q_{\psi}$) hidden layer & 3 \\
action value function ($Q_{\psi}$) hidden layer neuron & 256 \\
state value function ($V_{\phi}$) hidden layer & 3 \\
state value function ($V_{\phi}$) hidden layer neuron & 256 \\
intermediate energy function ($\mathcal{E}_{\Theta}$) hidden layer & 3 \\
intermediate energy function ($\mathcal{E}_{\Theta}$) hidden layer neuron & 256 \\
noise prediction function ($\epsilon_{\theta}$) hidden block & 7 \\
noise prediction function ($\epsilon_{\theta}$) hidden layer neuron & 256/512/1024 \\
inverse temperature $\beta$ & 1/2/3/4/5 \\
expectile weight $\tau$ & 0.5/0.6/0.7/0.8 \\
guidance degree $\omega$ & $0.0-3.0$ \\
$\nu$ & 0.001 \\
$\bar{a}$,$u_0$ & Refer to Appendix~\ref{taylor expansion fixed point} \\
\bottomrule
\end{tabular}
\end{table}

\subsection{Computation}\label{computation}
We conduct the experiments on NVIDIA GeForce RTX 3090 GPUs and NVIDIA A10 GPUs, and the CPU type is Intel(R) Xeon(R) Gold 6230 CPU @ 2.10GHz. 
Each run of the experiments spanned about 48-72 hours, depending on different tasks.

\subsection{Hyperparameters}
The hyperparameters used in our method are shown in Table~\ref{our method hyper}.

\section{The Constrained RL Problem}\label{The Constrained RL Problem}
The constrained RL problem is 
\begin{equation}
    \begin{aligned}
        \min_{\pi}~~&-\mathbb{E}_{s\sim D^\mu}\left[\mathbb{E}_{a\sim\pi(\cdot|s)}Q(s,a)\right]\\
        s.t.~~&D_{KL}(\pi(\cdot|s)||\mu(\cdot|s))\\
    \end{aligned},
\end{equation}
which can be converted to 
\begin{equation}\label{constrained RL problem lagrangian version}
    \max_{\pi}\mathbb{E}_{s\sim D^\mu}\left[\mathbb{E}_{a\sim\pi(\cdot|s)}Q(s,a)-\frac{1}{\beta}D_{KL}(\pi(\cdot|s)||\mu(\cdot|s))\right]. 
\end{equation}
It reveals the optimal solution
\begin{equation*}
    \pi^*(a|s)\propto\mu(a|s)e^{\beta Q(s,a)},
\end{equation*}
where $\pi^*(a|s)$ is the optimal policy that can produce optimal decisions by learning from the dataset, $\mu(a|s)$ is the behavior policy that is used to sample the data of the dataset, i.e., the data distribution of the dataset induced from $\mu(a|s)$, $Q(s, a)$ is the action value function, and we usually adopt neural network to learn the action value function.
\begin{proof}

The Equation~\eqref{constrained RL problem lagrangian version} is actually the lagrangian function of the contained RL problem by selecting the lagrangian multiplier as $\frac{1}{\beta}$ and adding the implicit constraint $\int \pi(a|s) da = 1$
\begin{equation*}
    \begin{aligned}
        &\min_{\pi}~~\mathbb{E}_{s\sim D^\mu}\mathcal{L}\\
        =&\min_{\pi}~~\mathbb{E}_{s\sim D^\mu}\left[\mathbb{E}_{a\sim\pi(\cdot|s)}Q(s,a)-\frac{1}{\beta}*D_{KL}(\pi(\cdot|s)||\mu(\cdot|s)) + \eta*(\int \pi(a|s) da - 1)\right]\\
    \end{aligned}.
\end{equation*}
The functional derivative of $\mathcal{L}$ regarding $\pi$ is
\begin{equation*}
    \begin{aligned}
        \frac{\partial \mathcal{L}}{\partial \pi}&=\frac{\mathcal{L}(\pi+\delta \pi) - \mathcal{L}(\pi)}{\partial \pi}\\
        \mathcal{L}(\pi+\delta \pi)&=\int (\pi(a|s) + \delta \pi(a|s)) Q_{\psi}(s,a) da\\
        &-\frac{1}{\beta}*\int (\pi(a|s) + \delta \pi(a|s))\log~\frac{\pi(a|s) + \delta \pi(a|s)}{\mu(\cdot|s)} da\\
        &+ \eta*\left(\int (\pi(a|s) + \delta \pi(a|s)) da - 1\right).\\
    \end{aligned}
\end{equation*}
Notice that the Taylor expansion of $\log~\frac{u(x)}{q(x)}$ at $u(x)=p(x)$ is 
\begin{equation*}
    \begin{aligned}
        \log~\frac{u(x)}{q(x)}|_{u(x)=p(x)}&=\log~\frac{p(x)}{q(x)}\\
        \left(\log~\frac{u(x)}{q(x)}\right)^{\prime}|_{u(x)=p(x)}&=\frac{1}{p(x)}\\
    \end{aligned}
\end{equation*}
Thus,
\begin{equation*}
    \begin{aligned}
        \log~\frac{u(x)}{q(x)} &\approx \log~\frac{u(x)}{q(x)}|_{u(x)=p(x)}+\left(\log~\frac{u(x)}{q(x)}\right)^{\prime}|_{u(x)=p(x)}(u(x) - p(x))
    \end{aligned}
\end{equation*}

Let $u(x)=\pi(a|s) + \delta \pi(a|s), p(x)=\pi(a|s), q(x)=\mu(a|s)$, we have 
\begin{equation*}
    \log~\frac{\pi(a|s) + \delta \pi(a|s)}{\mu(\cdot|s)}\approx \log~\frac{\pi(a|s)}{\mu(a|s)} + \frac{1}{\pi(a|s)}\delta \pi(a|s).
\end{equation*}

Then, $\mathcal{L}(\pi+\delta\pi)$ can be simplified 
\begin{equation*}
    \begin{aligned}
        \mathcal{L}(\pi+\delta\pi)&=\int (\pi(a|s) + \delta \pi(a|s)) Q(s,a) da\\
        &-\frac{1}{\beta}*\int (\pi(a|s) + \delta \pi(a|s))\left(\log~\frac{\pi(a|s)}{\mu(a|s)} + \frac{1}{\pi(a|s)}\delta \pi(a|s)\right) da\\
        &+ \eta*\left(\int (\pi(a|s) + \delta \pi(a|s)) da - 1\right).\\
        &=\int \pi(a|s) Q(s,a) da -\frac{1}{\beta}*\int \pi(a|s)\log~\frac{\pi(a|s)}{\mu(a|s)} da + \eta*\left(\int \pi(a|s) da - 1\right)\\
        &+\int \delta\pi(a|s)\left[Q(s,a)-\frac{1}{\beta}\left(\log~\frac{\pi(a|s)}{\mu(a|s)}+1\right)+\eta\right] -\frac{1}{\beta}\frac{1}{\pi(a|s)}(\delta\pi(a|s))^2\\
        &\approx \mathcal{L}(\pi) + \int \delta\pi(a|s)\left[Q(s,a)-\frac{1}{\beta}\left(\log~\frac{\pi(a|s)}{\mu(a|s)}+1\right)+\eta\right] da
    \end{aligned}
\end{equation*}
Finally, we obtain the functional derivative is 
\begin{equation*}
    \begin{aligned}
        \frac{\partial \mathcal{L}}{\partial \pi}&=\frac{\mathcal{L}(\pi+\delta \pi) - \mathcal{L}(\pi)}{\partial \pi}\\
        &=Q(s,a)-\frac{1}{\beta}\left(\log~\frac{\pi(a|s)}{\mu(a|s)}+1\right)+\eta
    \end{aligned}
\end{equation*}
Let $\frac{\partial\mathcal{L}}{\partial\pi}=0$, we have 
\begin{equation*}
    \begin{aligned}
        \pi^*(a|s) &= \mu(a|s)*e^{\beta(Q(s,a)+\eta)-1}\\
        \int \pi^*(a|s) da &= \int \mu(a|s)*e^{\beta(Q(s,a)+\eta)-1} da = 1\\
        \int \mu(a|s)*e^{\beta*Q(s,a)} da &= e^{\beta*\eta - 1}\\
        \pi^*(a|s) &= \frac{\mu(a|s)*e^{\beta*Q(s,a)}}{\int \mu(a|s)*e^{\beta*Q(s,a)} da}\\
        \pi^*(a|s) &\propto \mu(a|s)*e^{\beta*Q(s,a)}
    \end{aligned}
\end{equation*}
\end{proof}

\section{Analysis of Exact and Inexact Intermediate Guidance}\label{Exact Intermediate Guidance}

Previous studies propose that through an ingenious definition (For clarity, we rewrite it below), we can guarantee the relation $p_t(x_t)\propto q_t(x_t)e^{-\mathcal{E}_t(x_t)}$ of $p(x_t)$ and $q(x_t)$ at any time $t$, where $\mathcal{E}_t(x_t)$ is general formula of intermediate energy.

\begin{theorem}[Intermediate Energy Guidance]
\label{intermediate_energy_guidance}
Suppose $p_0(x_0)$ and $q_0(x_0)$ has the relation of Equation~\eqref{solution of energy-function-guided diffusion models}. For $t\in(0, T]$, let $p_{t|0}(x_t|x_0)$ and $q_{t|0}(x_t|x_0)$ be defined by
\begin{equation*}
    p_{t|0}(x_t|x_0) \coloneqq q_{t|0}(x_t|x_0) = \mathcal{N}(x_t;\alpha_t x_0,\sigma_t^2\bm{I}).
\end{equation*}
So the marginal distribution $p_t(x_t)$ and $q_t(x_t)$ at time $t$ are 
$p_t(x_t)=\int p_{t|0}(x_t|x_0) p_0(x_0)dx_0$ and $q_t(x_t)=\int q_{t|0}(x_t|x_0) q_0(x_0)dx_0$.
Define a general representation of intermediate energy as
\begin{equation}
    \mathcal{E}_t(x_t)=
    \begin{cases}
    \beta\mathcal{E}(x_0), & t = 0 \\
    -\log~\mathbb{E}_{q_{0|t}(x_0|x_t)}[e^{-\beta\mathcal{E}(x_0)}], & t > 0
    \end{cases}
\end{equation}
Then $q_t(x_t)$ and $p_t(x_t)$ satisfy
\begin{equation}
    p_t(x_t) \propto q_t(x_t)e^{-\mathcal{E}_t(x_t)}
\end{equation}
for any diffusion step $t$ and the score functions of $q_t(x_t)$ and $p_t(x_t)$ satisfy $\nabla_{x_t}\log p_t(x_t) = \nabla_{x_t} \log~q_t(x_t) - \nabla_x \mathcal{E}_t(x_t)$.
\end{theorem}
The derivation can be found in Appendix~\ref{Guidance of Intermediate Diffusion Steps}.

While for inexact intermediate energy 
\begin{equation*}
    \begin{aligned}
        \mathcal{E}^{MSE}_t(x_t)&=\mathbb{E}_{q_{0|t}(x_0|x_t)}[\mathcal{E}(x_0)], t>0,\\
        \mathcal{E}^{DPS}_t(x_t)&=\mathcal{E}(\mathbb{E}_{q_{0|t}(x_0|x_t)}[x_0]), t>0,
    \end{aligned}
\end{equation*}
that are adopted in classifier-guided and classifier-free-guided methods, it can be derivated that $\mathcal{E}_t(x_t)\geq\mathcal{E}^{MSE}_t(x_t)$ when $\beta=1$ and $t>0$.
\begin{proof}
By applying Jensen's Inequality, it is easy to have 
\begin{equation*}
    -\log~\mathbb{E}_{q_{0|t}(x_0|x_t)}[e^{-\beta\mathcal{E}(x_0)}]\geq-\mathbb{E}_{q_{0|t}(x_0|x_t)}[\log~e^{-\beta\mathcal{E}(x_0)}]=\beta\mathbb{E}_{q_{0|t}(x_0|x_t)}[\mathcal{E}(x_0)].
\end{equation*}
When $\beta=1$, we have $\mathcal{E}_t(x_t)\geq\mathcal{E}^{MSE}_t(x_t)$
\end{proof}

Also, we can prove that these three intermediate guidance values are not equal to each other: $\nabla_{x_t}\mathcal{E}_t(x_t)\neq\nabla_{x_t}\mathcal{E}^{MSE}_t(x_t)\neq\nabla_{x_t}\mathcal{E}^{DPS}_t(x_t)$.
\begin{proof}
For $t>0$, 
    \begin{equation*}
        \begin{aligned}
            \nabla_{x_t}\mathcal{E}_t(x_t)&=\nabla_{x_t}-\log~\mathbb{E}_{q_{0|t}(x_0|x_t)}[e^{-\beta\mathcal{E}(x_0)}],\\
            &=-\frac{1}{\mathbb{E}_{q_{0|t}(x_0|x_t)}[e^{-\beta\mathcal{E}(x_0)}]}\nabla_{x_t}\int q_{0|t}(x_0|x_t)[e^{-\beta\mathcal{E}(x_0)}]dx_0,\\
            &=-e^{\mathcal{E}_t(x_t)}\int q_{0|t}(x_0|x_t)\nabla_{x_t}\log q_{0|t}(x_0|x_t)e^{-\beta\mathcal{E}(x_0)}dx_0,\\
            &=-\mathbb{E}_{q_{0|t}(x_0|x_t)}\left[e^{\mathcal{E}_t(x_t)-\beta\mathcal{E}(x_0)}\nabla_{x_t}\log q_{0|t}(x_0|x_t)\right],
        \end{aligned}
    \end{equation*}
where we use the chain rule of derivation and gradient trick to obtain the result.
when $\beta=1$, we have $\nabla_{x_t}\mathcal{E}_t(x_t)=-\mathbb{E}_{q_{0|t}(x_0|x_t)}\left[e^{\mathcal{E}_t(x_t)-\mathcal{E}(x_0)}\nabla_{x_t}\log q_{0|t}(x_0|x_t)\right]$.
Similarly, we can obtain the gradient of $\mathcal{E}^{MSE}_t(x_t)$ and $\mathcal{E}^{DPS}_t(x_t)$:
\begin{equation*}
    \begin{aligned}
        \nabla_{x_t}\mathcal{E}^{MSE}_t(x_t)&=\nabla_{x_t}\mathbb{E}_{q_{0|t}(x_0|x_t)}[\mathcal{E}(x_0)],\\
        &=\nabla_{x_t}\int q_{0|t}(x_0|x_t)\mathcal{E}(x_0) dx_0,\\
        &=\int q_{0|t}(x_0|x_t)\nabla_{x_t}\log q_{0|t}(x_0|x_t)\mathcal{E}(x_0) dx_0,\\
        &=\mathbb{E}_{q_{0|t}(x_0|x_t)}\left[\mathcal{E}(x_0)\nabla_{x_t}\log q_{0|t}(x_0|x_t)\right].
    \end{aligned}
\end{equation*}
\begin{equation*}
    \begin{aligned}
        \nabla_{x_t}\mathcal{E}^{DPS}_t(x_t)&=\nabla_{x_t}\mathcal{E}(y)|_{y=\mathbb{E}_{q_{0|t}(x_0|x_t)}[x_0]},\\
        &=\nabla_{y}\mathcal{E}(y)^\top\nabla_{x_t}\mathbb{E}_{q_{0|t}(x_0|x_t)}[x_0],\\
        &=\nabla_{y}\mathcal{E}(y)^\top\mathbb{E}_{q_{0|t}(x_0|x_t)}\left[x_0\nabla_{x_t}\log q_{0|t}(x_0|x_t)\right],\\
        &=\mathbb{E}_{q_{0|t}(x_0|x_t)}\left[\nabla_{y}\mathcal{E}(y)^\top x_0\nabla_{x_t}\log q_{0|t}(x_0|x_t)\right].\\
    \end{aligned}
\end{equation*}
From the results we can see that $\nabla_{x_t}\mathcal{E}_t(x_t)\neq\nabla_{x_t}\mathcal{E}^{MSE}_t(x_t)\neq\nabla_{x_t}\mathcal{E}^{DPS}_t(x_t)$.
\end{proof}

\section{Guidance of Intermediate Diffusion Steps}\label{Guidance of Intermediate Diffusion Steps}

In order to guide the diffusion model during the intermediate diffusion steps, 
we first consider the following problem
\begin{equation}
    \begin{aligned}
    \min_{p}~~&\mathbb{E}_{x_0\sim p_0(x_0)}\mathcal{E}(x_0)\\
        s.t.~~&D_{KL}(p_0(x_0)||q_0(x_0))\\
    \end{aligned}
\end{equation}
and the relation function 
\begin{equation}\label{solution of general minimize problem}
    p_0(x_0)\propto q_0(x_0)e^{-\beta\mathcal{E}(x_0)}
\end{equation}
Define 
\begin{equation*}
    \begin{aligned}
        p_{t|0}(x_t|x_0) \coloneqq& q_{t|0}(x_t|x_0) = \mathcal{N}(x_t;\alpha_t x_0,\sigma_t^2\bm{I})\\
        p_t(x_t)=&\int p_{t|0}(x_t|x_0) p_0(x_0)dx_0,\\
        q_t(x_t)=&\int q_{t|0}(x_t|x_0) q_0(x_0)dx_0,\\
        \mathcal{E}_t(x_t)=&
    \begin{cases}
    \beta\mathcal{E}(x_0), & t = 0 \\
    -\log~\mathbb{E}_{q_{0|t}(x_0|x_t)}[e^{-\beta\mathcal{E}(x_0)}], & t > 0
    \end{cases}
    \end{aligned}
\end{equation*}
Then, we will obtain the following relation between score function of $p_t(x_t)$ and the score function of  of $q_t(x_t)$
\begin{equation}
    \nabla_{x_t} \log~p_t(x_t)=\nabla_{x_t} q_t(x_t)-\nabla_{x_t} \mathcal{E}_t(x_t)
\end{equation}
for each intermediate diffusion step.

\begin{proof}
Known that
\begin{equation*}
    \begin{aligned}
        & p_{t|0}(x_t|x_0)=q_{t|0}(x_t|x_0)=\mathcal{N}(x_t|\alpha_tx_0,\sigma_t^2\bm{I}) \\
        & q_t(x_t)=\int q_{t|0}(x_t|x_0)q_0(x_0)dx_0 \\
        & p_t(x_t)=\int p_{t|0}(x_t|x_0)p_0(x_0)dx_0 \\
        & p_0(x_0) \propto q_0(x_0)e^{-\beta \mathcal{E}(x_0)}\\
    \end{aligned}
\end{equation*}
By integral
\begin{equation*}
    Z=\int q_0(x_0)e^{-\beta\mathcal{E}(x_0)}dx_0=\mathbb{E}_{q_0(x_0)}\left[e^{-\beta\mathcal{E}(x_0)}\right],
\end{equation*}
and 
\begin{equation*}
    p_0(x_0) \propto q_0(x_0)e^{-\beta \mathcal{E}(x_0)},
\end{equation*}
we know 
\begin{equation*}
    p_0(x_0) = \frac{q_0(x_0)e^{-\beta \mathcal{E}(x_0)}}{Z}.
\end{equation*}
Thus,
\begin{equation*}
    \begin{aligned}
        p_t(x_t)&=\int p_{t|0}(x_t|x_0)p_0(x_0)dx_0\\
                &=\int p_{t|0}(x_t|x_0)\frac{q_0(x_0)e^{-\beta \mathcal{E}(x_0)}}{Z}dx_0\\
                &=\int q_{t|0}(x_t|x_0)\frac{q_0(x_0)e^{-\beta \mathcal{E}(x_0)}}{Z}dx_0\\
                &=q_t(x_t)\int q_{0|t}(x_0|x_t)\frac{e^{-\beta \mathcal{E}(x_0)}}{Z}dx_0\\
                &=\frac{q_t(x_t)\mathbb{E}_{q_{0|t}(x_0|x_t)}[e^{-\beta \mathcal{E}(x_0)}]}{Z}\\
                &=\frac{q_t(x_t)e^{-\mathcal{E}_t(x_t)}}{Z}\\
        p_t(x_t)&\propto q_t(x_t)e^{-\mathcal{E}_t(x_t)},
    \end{aligned}
\end{equation*}
\end{proof}
where we replace $q_{t|0}(x_t|x_0)q_0(x_0)$ with $q_{t}(x_t)q_{0|t}(x_0|x_t)$ by using Bayes Law in the fourth equation.
Now, if we want to generate data from $p(x_0)$, we just need train a diffusion model $q(x_0)$ and a guidance $\mathcal{E}_t$, then use 
\begin{equation}
    \begin{aligned}
        p_t(x_t)&\propto q_t(x_t)e^{-\mathcal{E}_t(x_t)}\\
        \nabla_{x_t}\log~p_t(x_t)&= \nabla_{x_t}\log~q_t(x_t)-\nabla_{x_t}\mathcal{E}_t(x_t)
    \end{aligned}
\end{equation}
to obtain the score function value of $\nabla_{x_t}\log~p_t(x_t)$.
Then, we can use the reverse transformation (Equation~\eqref{neural ODE}) to generate samples.

In RL, we should use $Q(s,a)$ to denote $-\mathcal{E}(x_0)$, because Equation~\eqref{energy-function-guided diffusion models} and Equation~\eqref{constrained RL problem} indicate that maximizing $Q(s,a)$ is same with minimizing $\mathcal{E}_0(x_0)$.
Besides, we notice that approximating $\mathcal{E}_t(x_t)$ needs to calculate $\mathcal{E}(x_0)$.
And actually $\mathcal{E}(x_0)$ is the $Q(s,a)$.
In other words, the intermediate energy guidance in RL is defined by
\begin{equation}
    \mathcal{E}_t(x_t)=
    \begin{cases}
    \beta\mathcal{E}(x_0)\rightarrow\beta Q(s,a), & t = 0 \\
    -\log~\mathbb{E}_{q_{0|t}(x_0|x_t)}[e^{-\beta\mathcal{E}(x_0)}]\rightarrow \log~\mathbb{E}_{q_{0|t}(x_0|x_t)}[e^{\beta Q(s,a)}], & t > 0
    \end{cases}
\end{equation}
\begin{proof}
\begin{equation*}
    \begin{aligned}
        \pi_t(a_t|s)&=\int \pi_{t|0}(a_t|a_0,s)\pi_0(a_0|s)da_0\\
                &=\int \pi_{t|0}(a_t|a_0,s)\frac{\mu_0(a_0|s)e^{\beta Q(s,a_0)}}{Z}da_0\\
                &=\int \pi_{t|0}(a_t|a_0,s)\frac{\mu_0(a_0|s)e^{\beta Q(s,a_0)}}{Z}da_0\\
                &=\mu_t(a_t|s)\int \mu_{0}(a_0|a_t,s)\frac{e^{\beta Q(s,a_0)}}{Z}da_0\\
                &=\frac{\mu_t(a_t|s)\mathbb{E}_{\mu_{0}(a_0|a_t,s)}[e^{\beta Q(s,a_0)}]}{Z}\\
                &=\frac{\mu_t(a_t|s)e^{\mathcal{E}_t(a_t)}}{Z}\\
        \pi_t(a_t|s)&\propto \mu_t(a_t|s)e^{\mathcal{E}_t(s,a_t)}
    \end{aligned}
\end{equation*}
\end{proof}
Here, we slightly abuse the notion of $\mathcal{E}$ because, in RL, the input of the energy function is the concatenation of state and action.

\section{Detailed Derivation of Intermediate Guidance}\label{Detailed Derivation of Intermediate Guidance}

Implicit dependence of $Q(s, a_0)$ on action $a_0$ hinds the exact calculation of $\log~\mathbb{E}_{\mu_{0|t}(a_0|a_t,s)}[e^{\beta Q(s,a_0)}]$.
In order to approximate the intermediate energy $\mathcal{E}_t(s, a_t)$ shown in Equation~\eqref{intermediate energy definition in RL}, we first expand $Q(s, a_0)$ at $a_0=\bar{a}$ with Taylor expansion, i.e.,
\begin{equation*}
    \begin{aligned}
        Q(s, a_0) &\approx Q(s, a_0)|_{a_0=\bar{a}} + \frac{\partial Q(s, a_0)}{\partial a_0}|_{a_0=\bar{a}}*(a_0-\bar{a}),
    \end{aligned}
\end{equation*}
where $\bar{a}$ is a constant vector.
Replacing it to $\log~\mathbb{E}_{a_{0}\sim \mu(a_0|a_t,s)}e^{\beta Q(s, a_0)}$, we can transfer the implicit dependence of $Q(s, a_0)$ on $a_0$ to explicit dependence on $Q^{\prime}(s, \bar{a})*a_0$.
Accordingly, $\log~\mathbb{E}_{a_{0}\sim \mu(a_0|a_t,s)}e^{\beta Q(s, a_0)}$ is given by
\begin{equation*}
    \begin{aligned}
        &\log~\mathbb{E}_{a_{0}\sim \mu(a_0|a_t,s)}e^{\beta Q(s, a_0)}\\
        \approx&\log~\mathbb{E}_{a_{0}\sim \mu(a_0|a_t,s)}e^{\beta*\left(Q(s, a_0)|_{a_0=\bar{a}} + \frac{\partial Q(s, a_0)}{\partial a_0}|_{a_0=\bar{a}}*(a_0-\bar{a})\right)}\\
        =&\log~\mathbb{E}_{a_{0}\sim \mu(a_0|a_t,s)}e^{\beta*\left(Q(s, \bar{a}) + Q^{\prime}(s, \bar{a})*(a_0-\bar{a})\right)}\\
        =&\log~\mathbb{E}_{a_{0}\sim \mu(a_0|a_t,s)}\left[e^{\beta*Q(s, \bar{a})}*e^{\beta*Q^{\prime}(s, \bar{a})*(a_0-\bar{a})}\right]\\
        =&\log~\left\{e^{\beta*Q(s, \bar{a})-\beta*Q^{\prime}(s,\bar{a})*\bar{a}}\right\}+\log~\left\{\mathbb{E}_{a_{0}\sim \mu(a_0|a_t,s)}[e^{\beta*Q^{\prime}(s, \bar{a})*a_0}]\right\}\\
        =&\beta*Q(s, \bar{a})-\beta*Q^{\prime}(s,\bar{a})*\bar{a}+\log~\left\{\mathbb{E}_{a_{0}\sim \mu(a_0|a_t,s)}[e^{\beta*Q^{\prime}(s, \bar{a})*a_0}]\right\}.
    \end{aligned}
\end{equation*}
Furthermore, by applying the moment generating function of Gaussian distribution, we have
\begin{equation*}
    \begin{aligned}
        &\log~\mathbb{E}_{a_{0}\sim \mu(a_0|a_t,s)}e^{\beta Q(s, a_0)}\\
        &=\beta*Q(s, \bar{a})-\beta*Q^{\prime}(s,\bar{a})*\bar{a}+\log~\left\{\mathbb{E}_{a_{0}\sim \mu(a_0|a_t,s)}[e^{\beta*Q^{\prime}(s, \bar{a})*a_0}]\right\}\\
        &=\beta*Q(s, \bar{a})-\beta*Q^{\prime}(s,\bar{a})*\bar{a}+\beta*Q^{\prime}(s, \bar{a})^\top*\tilde{\mu}_{0|t}+\frac{1}{2}*\beta^2*\tilde{\sigma}_{0|t}^2*||Q^{\prime}(s, \bar{a})||^2_2\\
        &=\beta*Q(s, \bar{a})+\beta*Q^{\prime}(s, \bar{a})^\top*(\tilde{\mu}_{0|t}-\bar{a})+\frac{1}{2}*\beta^2*\tilde{\sigma}_{0|t}^2*||Q^{\prime}(s, \bar{a})||^2_2.\\
    \end{aligned}
\end{equation*}

\section{Detailed Derivation of Posterior Distribution}

\subsection{Posterior 1}\label{Posterior 1 of appendix}
If we want to get the exact mean of $\mu_{0|t}(a_0|a_t,s)$, we actually want to optimize the following objective~\cite{bao2022analytic}
\begin{equation}
    \tilde{\mu}^*_{0|t}(a_0|a_t,s)=\min_{\theta} \mathbb{E}_{a_0\sim\mu_{0|t}(a_0|a_t,s)}\left[||\tilde{\mu}_{0|t,\theta}(a_0|a_t,s)-a_0||^2_2\right],
\end{equation}
where $\tilde{\mu}_{0|t,\theta}(a_0|a_t,s)$ can be further reparameterized by 
\begin{equation}\label{reverse reparameterization}
    \tilde{\mu}_{0|t,\theta}=\frac{1}{\alpha_t}(a_t-\sigma_t\epsilon_\theta(s,a_t,t)),
\end{equation}
because this reparameterization follows the same loss function (as shown in Equation~\eqref{general diffusion loss}) as the diffusion model, and we do not need to introduce additional parameters to approximate the mean of $\mu_{0|t}(a_0|a_t,s)$.
For the covariance matrix $\tilde{\Sigma}_{0|t}$, following the definition of covariance
\begin{equation}
    \begin{aligned}
        \tilde{\Sigma}_{0|t}(a_t) &= \mathbb{E}_{\mu_{0|t}(a_0|a_t,s)}\left[(a_0 - \tilde{\mu}_{0|t})(a_0 - \tilde{\mu}_{0|t})^\top\right]\\
        &=\mathbb{E}_{\mu_{0|t}(a_0|a_t,s)}\left[(a_0 - \frac{1}{\alpha_t}(a_t-\sigma_t\epsilon_\theta))(a_0 - \frac{1}{\alpha_t}(a_t-\sigma_t\epsilon_\theta))^\top\right]\\
        &=\mathbb{E}_{\mu_{0|t}(a_0|a_t,s)}\left[(a_0 - \frac{1}{\alpha_t}a_t)(a_0 - \frac{1}{\alpha_t}a_t)^\top\right]-\frac{\sigma_t^2}{\alpha_t^2}\epsilon_\theta\epsilon_\theta^\top\\
        &=\frac{1}{\alpha_t^2}\mathbb{E}_{\mu_{0|t}(a_0|a_t,s)}\left[(a_t - \alpha_t a_0)(a_t - \alpha_t a_0)^\top\right]-\frac{\sigma_t^2}{\alpha_t^2}\epsilon_\theta\epsilon_\theta^\top,
    \end{aligned}
\end{equation}
where in the second equation we replace $\tilde{\mu}_{0|t}$ with Equation~\eqref{reverse reparameterization}.
In the third equation we use
\begin{equation*}
    \begin{aligned}
        &\mathbb{E}_{\mu_{0|t}(a_0|a_t)}[(a_0 - \frac{1}{\alpha_t}a_t)*\frac{\sigma_t}{\alpha_t}\epsilon_\theta]\\
        &=(\mathbb{E}_{\mu_{0|t}(a_0|a_t,s)}[a_0] - \frac{1}{\alpha_t}a_t)*\frac{\sigma_t}{\alpha_t}\epsilon_\theta\\
        &=(\tilde{\mu}_{0|t} - \frac{1}{\alpha_t}a_t)*\frac{\sigma_t}{\alpha_t}\epsilon_\theta\\
        &=-\frac{\sigma_t^2}{\alpha_t^2}\epsilon_\theta\epsilon_\theta^T,
    \end{aligned}
\end{equation*}
and $\tilde{\mu}_{0|t}=\mathbb{E}_{\mu_{0|t}(a_0|a_t,s)}[a_0]$.
Besides, we also notice that $\mu(a_t)\sim\mathcal{N}(a_t;\alpha_t a_0,\sigma_t^2\bm{I})$, thus we have
\begin{equation*}
    \begin{aligned}
        &\mathbb{E}_{\mu_t(a_t|s)}\mathbb{E}_{\mu_{0|t}(a_0|a_t,s)}\left[(a_t - \alpha_t a_0)(a_t - \alpha_t a_0)^\top\right]\\
        =&\mathbb{E}_{\mu(a_0|s)}\mathbb{E}_{\mu_{t|0}(a_t|a_0,s)}\left[(a_t - \alpha_t a_0)(a_t - \alpha_t a_0)^\top\right]\\
        =&\mathbb{E}_{\mu(a_0|s)}\mathbb{E}_{\mu_{t|0}(a_t|a_0,s)}\left[(a_t - \alpha_t a_0)(a_t - \alpha_t a_0)^\top\right]\\
        =&\mathbb{E}_{\mu(a_0|s)}\Sigma_{t|0}\\
        =&\sigma_t^2\bm{I},
    \end{aligned}
\end{equation*}
where in the first equation $\mu_t(a_t|s)\mu_{0|t}(a_0|a_t,s)=\mu(a_0|s)\mu_{t|0}(a_t|a_0,s)$.
To make the variance independent from the data $a_t$, we can apply the expectation over $\tilde{\Sigma}_{0|t}(a_t)$:
\begin{equation*}
    \begin{aligned}
        \tilde{\Sigma}_{0|t}&=\mathbb{E}_{\mu_t(a_t|s)}\tilde{\Sigma}_{0|t}(a_t)\\
        &=\frac{\sigma_t^2}{\alpha^2_t}[\bm{I}-\mathbb{E}_{\mu_t(a_t|s)}[\epsilon_\theta(a_t, t)\epsilon_\theta(a_t, t)^\top]].
    \end{aligned}
\end{equation*}
For simplicity, we usually consider isotropic Gaussian distribution, i.e., $\tilde{\Sigma}_{0|t}=\tilde{\sigma}^2_{0|t}\bm{I}$, which indicates that
\begin{equation*}
    \tilde{\sigma}^2_{0|t} = \frac{\sigma_t^2}{\alpha^2_t}[1-\frac{1}{d}\mathbb{E}_{\mu_t(a_t|s)}[||\epsilon_\theta(a_t, t)||^2_2]].
\end{equation*}

\subsection{Posterior 2}\label{Posterior 2 of appendix}

For the mean value of the distribution $\mu_{0|t}(a_0|a_t,s)$, we still adopt the same reparameterization trick $\tilde{\mu}_{0|t,\theta}=\frac{1}{\alpha_t}(a_t-\sigma_t\epsilon_\theta(s,a_t,t))$.
But for the covariance, we can reformulate~\cite{kexuefm9246} it as 
\begin{equation}
    \begin{aligned}
        \tilde{\Sigma}_{0|t}(a_t)&= \mathbb{E}_{\mu_{0|t}(a_0|a_t,s)}\left[(a_0 - \tilde{\mu}_{0|t})(a_0 - \tilde{\mu}_{0|t})^\top\right]\\
        &=\mathbb{E}_{\mu_{0|t}(a_0|a_t,s)}\left[\left((a_0 - u_0)-(\tilde{\mu}_{0|t} - u_0)\right)\left((a_0 - u_0)-(\tilde{\mu}_{0|t} - u_0)\right)^\top\right]\\
        &=\mathbb{E}_{\mu_{0|t}(a_0|a_t,s)}\left[(a_0 - u_0)(a_0 - u_0)^\top\right]-(\tilde{\mu}_{0|t} - u_0)(\mu_{0|t} - u_0)^\top,
    \end{aligned}
\end{equation}
where we add a constant vector $u_0$ that has the same dimension with $\tilde{\mu}_{0|t}$ in the second equation.
We use $\tilde{\mu}_{0|t}=\mathbb{E}_{\mu_{0|t}(a_0|a_t,s)}[a_0]$ to derivate the third equation.
Perform expectation on $\tilde{\Sigma}_{0|t}(a_t)$ and let $u_0=\mathbb{E}_{a_0\sim\mu(a_0)}[a_0]$, we have
\begin{equation}
    \begin{aligned}
        \bar{\sigma}_t^2&=\mathbb{E}_{\mu_t(a_t|s)}[\tilde{\Sigma}_{0|t}(a_t)], \\ &= \mathbb{E}_{\mu_t(a_t|s)}\mathbb{E}_{\mu_{0|t}(a_0|a_t,s)}\left[(a_0 - u_0)(a_0 - u_0)^\top\right]-\mathbb{E}_{\mu_t(a_t|s)}\left[(\tilde{\mu}_{0|t} - u_0)(\tilde{\mu}_{0|t} - \mu_0)^\top\right],\\
        &=\mathbb{E}_{\mu(a_0|s)}\mathbb{E}_{\mu_{t|0}(a_t|a_0,s)}\left[(a_0 - u_0)(a_0 - u_0)^\top\right]-\mathbb{E}_{\mu_t(a_t|s)}(\tilde{\mu}_{0|t} - u_0)(\tilde{\mu}_{0|t} - u_0)^\top,\\
        &=\frac{1}{d}\mathbb{E}_{\mu(a_0|s)}\left[||a_0 - u_0||^2_2\right]-\frac{1}{d}\mathbb{E}_{\mu_{t}(a_t|s)}[||\tilde{\mu}_{0|t} - u_0||^2_2],\\
        &=Var(a_0) - \frac{1}{d}\mathbb{E}_{\mu_t(a_t|s)}[||\tilde{\mu}_{0|t} - u_0||^2_2],
    \end{aligned}
\end{equation}
where $Var(a_0)$ can be approximated from a batch of data or from the entire dataset and $\mathbb{E}_{\mu_t(a_t|s)}[||\tilde{\mu}_{0|t} - u_0||^2_2]$ can be estimated by sampling a batch of data $a_0$ and $a_t$ with different $t$.

\subsection{Another Method to Approximate Posterior $\mu_{0|t}$}
Consider the covariance estimation under  $\mu_{0|t}\neq\mathbb{E}_{\mu_{0|t}(a_0|a_t,s)}[a_0]$ where $\mu_{0|t}$ has been trained well using the diffusion loss function (shown in Equation~\eqref{general diffusion loss}), we can adopt the following negative log-likelihood optimization problem~\cite{bao2022estimating}
\begin{equation*}
    \min_{\tilde{\Sigma}_{0|t}} \mathbb{E}_{a_t\sim \mu_t(a_t|s),a_0\sim \mu_{0|t}(a_0|a_t,s)}[-\log P(a_0; \tilde{\mu}_{0|t}, \tilde{\Sigma}_{0|t})].
\end{equation*}
perform derivation w.r.t $\tilde{\Sigma}_{0|t}$, we have
\begin{equation*}
    \begin{aligned}
        &\frac{\partial}{\partial \tilde{\Sigma}_{0|t}}-\log~P(a_0; \tilde{\mu}_{0|t}, \tilde{\Sigma}_{0|t}),\\
        &=\frac{\partial}{\partial \tilde{\Sigma}_{0|t}}-\log~\left[\frac{1}{{2\pi}^{d/2}|\tilde{\Sigma}_{0|t}|^{1/2}}e^{-\frac{1}{2}(a_0-\tilde{\mu}_{0|t})^{\top}\tilde{\Sigma}_{0|t}^{-1}(a_0-\tilde{\mu}_{0|t})}\right],\\
        &=\frac{\partial}{\partial \tilde{\Sigma}_{0|t}}\left[\frac{d}{2}\log~2\pi+\frac{1}{2}\log|\tilde{\Sigma}_{0|t}|+\frac{1}{2}(a_0-\tilde{\mu}_{0|t})^{\top}\tilde{\Sigma}_{0|t}^{-1}(a_0-\tilde{\mu}_{0|t})\right],\\
        &=\frac{1}{2}\tilde{\Sigma}_{0|t}^{-1}-\frac{1}{2}\tilde{\Sigma}_{0|t}^{-1}(a_0-\tilde{\mu}_{0|t})(a_0-\tilde{\mu}_{0|t})^{\top}\tilde{\Sigma}_{0|t}^{-1}.
    \end{aligned}
\end{equation*}
So the objective reaches maximization when $\tilde{\Sigma}_{0|t}^{*}=\mathbb{E}_{a_t\sim \mu_t(a_t|s),a_0\sim \mu_{0|t}(a_0|a_t,s)}\left[(a_0-\tilde{\mu}_{0|t})(a_0-\tilde{\mu}_{0|t})^{\top}\right]$.
Usually, we consider the isotropic Gaussian distribution, i.e., $\tilde{\Sigma}_{0|t}=\tilde{\sigma}_{0|t}^2\bm{I}$ and 
\begin{equation}\label{variance of posterior 3}
    \tilde{\sigma}_{0|t}^2=\frac{1}{d}\mathbb{E}_{a_0\sim \mu(a_0|s),a_t\sim \mu_{t|0}(a_t|a_0,s)}[||a_0-\tilde{\mu}_{0|t}||^2_2].
\end{equation}
Furthermore, we also have the following result, which is the same as Equation~\eqref{variance of posterior 3}:
\begin{equation}
    \begin{aligned}
        \tilde{\sigma}_{0|t}^2&=\frac{1}{d}\mathbb{E}_{a_0\sim \mu(a_0|s),a_t\sim \mu_{t|0}(a_t|a_0,s)}[||a_0-\tilde{\mu}_{0|t}||^2_2],\\
        &=\frac{1}{d}\mathbb{E}_{a_0\sim \mu(a_0|s), \epsilon\sim\mathcal{N}(0,\bm{I}), t\sim U[0,T], a_t=\alpha_t a_0+\sigma_t\epsilon}\left[||a_0-\frac{1}{\alpha_t}(a_t-\sigma_t\epsilon_\theta(s,a_t,t))||^2_2\right],\\
        &=\frac{\sigma_t^2}{\alpha_t^2 d}\mathbb{E}_{a_0\sim \mu(a_0|s), \epsilon\sim\mathcal{N}(0,\bm{I}), t\sim U[0,T], a_t=\alpha_t a_0+\sigma_t\epsilon}\left[||\epsilon-\epsilon_\theta(s,a_t,t)||^2_2\right].\\
    \end{aligned}
\end{equation}

Inspired by the previous studies~\cite{bao2022estimating, lu2022dpmb} that $\tilde{\mu}_{0|t}=\mathbb{E}_{\mu_{0|t}(a_0|a_t,s)}[a_0]$ is suitable for training in practice, so we only focus on the performance of Posterior 1 and Posterior 2 in this paper.

We summarize the posterior mean and covariance in
\begin{equation}\label{mean and variance solution of posterior}
    \begin{aligned}
        \tilde{\mu}_{0|t}&=\frac{1}{\alpha_t}(a_t-\sigma_t\epsilon_\theta(s,a_t,t)),\\
        \tilde{\sigma}_{0|t}^2 &= \frac{\sigma_t^2}{\alpha^2_t}\left[1-\frac{1}{d}\mathbb{E}_{\mu_t(a_t|s)}\left[||\epsilon_\theta(a_t, t)||^2_2\right]\right],\\
        \tilde{\sigma}_{0|t}^2 &= Var(a_0) - \frac{1}{d}\mathbb{E}_{\mu_t(a_t)}[||\tilde{\mu}_{0|t} - u_0||^2_2],\\
        \tilde{\sigma}_{0|t}^2&=\frac{\sigma_t^2}{\alpha_t^2 d}\mathbb{E}_{a_0\sim \mu(a_0|s), \epsilon\sim\mathcal{N}(0,\bm{I}), t\sim U[0,T], a_t=\alpha_t a_0+\sigma_t\epsilon}\left[||\epsilon-\epsilon_\theta(s,a_t,t)||^2_2\right].
    \end{aligned}
\end{equation}

\section{The Training of Q Function}\label{The Training of Q Function}

\subsection{Q Function Training with Expectile Regression}
Expectile regression loss 
\begin{equation*}
    \begin{aligned}
        \mathcal{L}_{V}&=\mathbb{E}_{(s,a)\sim D_\mu}\left[L_2^\tau(V_{\phi}(s)-Q_{\bar{\psi}}(s,a))\right],\\
        \mathcal{L}_{Q}&=\mathbb{E}_{(s,a,s^\prime)\sim D_\mu}\left[||r(s,a) + V_{\phi}(s^\prime)-Q_{\psi}(s,a)||^2_2\right],\\
        L_2^\tau(y)&=|\tau-1(y<0)|y^2,
    \end{aligned}
\end{equation*}
provides a method to avoid out-of-sample actions entirely and simultaneously perform maximization over the actions implicitly.
The target value of Q is induced from a parameterized state value function, which is trained by expectile regression on in-sample actions.
Besides, its effectiveness has been verified by many recent studies~\cite{kostrikov2021offline, hansen2023idql}.
Thus, we adopt this method to train the Q function.

\subsection{Q Function Training with In-support Softmax Q-Learning}

Due to the natural property of data augmentation in generative models, such as diffusion models, we can use diffusion models to generate fake samples~\cite{lu2023contrastive} and augment the training of the Q function:
\begin{equation*}
    \begin{aligned}
        \mathcal{L}_{ISQL}&=\mathbb{E}_{(s,a,s')\sim D}\left[||Q_{\psi}(s,a)-\mathcal{T}^{\pi}Q_{\psi}(s,a)||^2_2\right],\\
        \mathcal{T}^{\pi}Q_{\psi}(s,a)&\approx r(s,a)+\gamma*\frac{\sum_{\hat{a}^{\prime}}\left[e^{\beta Q_{\psi}(s^{\prime},\hat{a}^{\prime})}*Q_{\psi}(s^{\prime},\hat{a}^{\prime})\right]}{\sum_{\hat{a}^{\prime}}e^{\beta Q_{\psi}(s^{\prime},\hat{a}^{\prime})}},
    \end{aligned}
\end{equation*}
where $\hat{a}^{\prime}$ is the fake actions. 
This method needs additional samples to train the Q function, where we can not make sure all the fake actions are reasonable for certain states.
To a certain extent, these fake actions are OOD actions for the value functions.
Besides, additional actions will also cause large memory consumption.

\subsection{Q Function Training with Contrastive Q-Learning}

Contrastive Q-learning~\cite{kumar2020conservative} is another notable method for training the Q function from offline datasets.
It underestimates the action values for all actions that do not exist in the dataset to reduce the influence of OOD actions.
The training loss is 
\begin{equation*}
    \begin{aligned}
        \mathcal{L}_{CQL}&=\mathbb{E}_{s,a,s'\sim D}\left[||Q_{\psi}(s,a)-(r+Q_{\bar{\psi}}(s',a'=\pi(a'|s')))||^2_2\right]\\
        &+\lambda\left(\mathbb{E}_{s\sim D, a\sim\pi(a|s)}[Q_{\psi}(s,a)]-\mathbb{E}_{(s,a)\sim D}[Q_{\psi}(s,a)]\right).
    \end{aligned}
\end{equation*}
The learned Q function tends to predict similar values for actions that do not exist in datasets, which will cause ineffective intermediate guidance and further affect the performance.

\section{Implementation Details}

\subsection{Guidance Rescale Strategy}
The guidance scale affects the amplitude and direction of the generation.
The direction of generation must be changed due to the gradient guidance of the critic.
So we can 
We find that when the guidance scale is zero, the inference performance is more stable than the guidance scale is non-zero, which inspires us to modify the amplitude of the score of the optimal data distribution $\nabla_{a_t}\log~\pi_t(a_t|s)$.
Mathematically, we re-normalize the amplitude of $\nabla_{a_t}\log~\pi_t(a_t|s)$ by 
\begin{equation}
    \nabla_{a_t}\log~\pi_t(a_t|s) = \frac{\nabla_{a_t}\log~\pi_t(a_t|s)}{||\nabla_{a_t}\log~\pi_t(a_t|s)||}*||\nabla_{a_t}\log~\mu_t(a_t|s)||.
\end{equation}

\subsection{The Choice of $\bar{a}$ and $u_0$} \label{taylor expansion fixed point}

Although we can choose $\bar{a}$ as any vectors, considering the error bound of Taylor expansion and the stability of the training process, we use the following method to obtain $\bar{a}$
\begin{equation}\label{calculate bar_a}
    \begin{aligned}
        \bar{a} &= a - \nu*\frac{Q^\prime_{\psi}(s, a_0)}{||Q^\prime_{\psi}(s, a_0)||},\\
    \end{aligned}
\end{equation}
where $\nu=0.001$.
Considering the loss term of second-order approximation
\begin{equation*}
    R_2(a) = \frac{1}{2}\nabla_a^2 Q(s, c)(a-\bar{a})^2,
\end{equation*}
where $c$ is the point between $[\bar{a}, a]$. 
Considering the definition of Equation~\eqref{calculate bar_a},
\begin{equation*}
    \begin{aligned}
        &||\nabla_a Q(s, a)||*||a-\bar{a}||\\
        =&||\nabla_a Q(s, a)||*\nu*||\frac{Q^\prime_{\psi}(s, a_0)}{||Q^\prime_{\psi}(s, a_0)||}||\\
        =&\nu.
    \end{aligned}
\end{equation*}
So the error satisfies 
\begin{equation*}
    R_2(a) = \frac{1}{2}\left[(\nabla_a Q(s, c))^T (a-\bar{a})\right]^2 \approx \frac{1}{2}\left[(\nabla_a Q(s, a))^T (a-\bar{a})\right]^2 \leq \frac{1}{2} \nu^2
\end{equation*}
as long as $\bar{a}$ is close to $a$, i.e., the derivative between $\bar{a}$ and $a$ is approximately constant.
In practice, we use $Q^\prime(s, a)$ as the value of $Q^\prime(s, \bar{a})$ for two reasons: (1) Small $\nu$ value leads to approximately same derivative value at $a$ and $\bar{a}$. (2) Directly use $Q^\prime(s, a)$ will reduce one calculation of gradient for $Q^\prime(s,\bar{a})$.

In Equation~\eqref{mean and variance solution of posterior}, we approximate $\mathbb{E}_{q(a_t)}[||\epsilon_\theta(a_t, t)||^2_2]$ by sampling a batch of data and the calculation process is 
(1) Sample a batch of data $a_0$ from the dataset and sample $\epsilon$ from Gaussian distribution $\mathcal{N}(0, \bm{I})$.
(2) Sample time step index $t$.
(3) Obtain $a_t=\alpha_t a_0 + \sigma_t \epsilon$.
(4) Get the noise prediction norm with diffusion model $\epsilon_\theta(a_t, t, s)$.
(5) Average on the batch of data.

In Equation~\eqref{mean and variance solution of posterior}, $u_0$ serves as a constant baseline to calculate $\bar{\sigma}_t^2$, where $\mathbb{E}_{\mu(a_t)}[||\tilde{\mu}_{0|t} - u_0||^2_2]$ can be approximated from the data.
(1) We can sample a batch size of data $a_0$ from the dataset and sample $\epsilon$ from Gaussian distribution $\mathcal{N}(0, \bm{I})$. 
(2) Sample time step index $t$.
(3) Obtain $a_t=\alpha_t a_0 + \sigma_t \epsilon$.
(4) Use the diffusion model to predict $\mu_{0|t}=\frac{1}{\alpha_t}(a_t-\sigma_t\epsilon_\theta(a_t, t, s))$.
(5) Calculate $\mathbb{E}_{\mu(a_t)}[||\tilde{\mu}_{0|t} - u_0||^2_2]$ and average on the batch data.

\end{document}